\documentclass[review,10pt]{JMtemplate}
\usepackage{bm}
\usepackage{graphicx} 
\usepackage{amsmath,mathrsfs,amssymb} 
\usepackage{amsfonts}  
\usepackage{algorithm}
\usepackage{algorithmic}
\usepackage{siunitx}   
\usepackage{upgreek}   
\usepackage{xcolor}    
\usepackage{subfigure}
\usepackage{booktabs}      
\usepackage{longtable}
\usepackage{threeparttable}
\usepackage{multirow}
\usepackage{multicol}
\usepackage{times}
\usepackage{helvet}
\usepackage{courier}
\usepackage{marvosym}
\usepackage[hyphens]{url} 
\usepackage{natbib}  
\usepackage{caption} 
\usepackage{adjustbox}
\usepackage{subcaption}
\usepackage{subfigure}
\usepackage{makecell}

\newtheorem{theorem}{Theorem}

\newcommand*{\dif}{\mathop{}\!\mathrm{d}}

\newcommand*{\esssup}{\operatorname*{ess\,sup}}

\usepackage[
    colorlinks=true,    
    linkcolor=blue,     
    urlcolor=blue,      
    citecolor=green,    
    filecolor=magenta   
]{hyperref}
\usepackage{textcomp}

\begin{document}
\begin{frontmatter}
\title{On the Expressive Power of Weight Quantization\\ in Large Language Models}

\author{\textbf{Shao-Qun Zhang}\textsuperscript{1,2}  \\[0.3em]
\small\textsuperscript{1} National Key Laboratory for Novel Software Technology, Nanjing University, Nanjing 210063, China.\\
\small\textsuperscript{2} School of Intelligent Science and Technology, Nanjing University, Suzhou 215163, China.\\
\texttt{zhangsq@lamda.nju.edu.cn}
}

\begin{abstract}
In recent years, weight quantization that encodes the learnable parameters of large language models in an $n$-bit format has garnered significant attention due to its potential for model compression and inference acceleration. Many practical techniques have been developed; however, the theoretical understanding of many aspects, especially the approximation and degradation of expressive power as the number of quantization bits decreases, remains unclear. In this paper, we provide a theoretical investigation into the expressive capability of large language models relative to the number of quantization bits. We argue that 1.58-bit is the limiting precision for weight quantization by establishing the universal approximation and expressive collapse properties of weight-quantized models with respect to the number of quantization bits. Additionally, we confirm that weight quantization leads to expressive degradation, in which the expressive capacity of weight-quantized models degrades polynomially as the number of quantization bits decreases. These theoretical findings provide a solid foundation for advancing weight quantization in the context of scaling laws and shed insights for future research in model compression and inference acceleration.

\end{abstract}
\end{frontmatter}

\section{Introduction}\label{sec:introduction}
Weight quantization is a key technique encountered in lightweight applications of deep learning and Large Language Models (LLMs) to resource-constrained environments~\citep{gope2019ternary,zhang2025:ternaryclip}. Recent years have witnessed great efforts that weight quantization algorithms achieve model compression~\citep{li2016ternary,gholami2021:generalization}, inference acceleration~\citep{courbariaux2015binaryconnect,shen2024agile}, and efficient scaling~\citep{ouyang2024:wq,ma2025:wq}, while maintaining comparable accuracy to full-precision counterparts~\citep{zhou2017incremental,wu2025bitnet}. Some developers are even interested in attaining strong performance on particular downstream tasks with minimal computational overhead~\citep{wu2025bitnet,zhang2025:ternaryclip}. Despite practical progress, theoretical understandings of the expressive capacity of weight-quantized LLMs with various quantization bits are far from clear.

One important characterization is the universal approximation property of LLMs relative to various quantization bits, which provides a fundamental guarantee for weight quantization. In traditional deep learning theory, the universal approximation properties of various models, such as the feed-forward neural network~\citep{hornik1989:UA,gonon2023:UA}, spiking neural network~\citep{zhang2022:SNNsTheory}, complex-valued neural network~\citep{voigtlaender2023:UA}, and Transformers~\citep{yun2019:transformers,cheng2025:transformer}, equipped with real-valued or floating weights, have been built. Currently, the theoretical investigations on the approximation universality of LLMs with quantized weights are still limited. \citet{yayla2021:ua} investigated the universal approximation of 1-bit neural networks with binary and real-valued inputs. \citet{ding2019universal} proved that extremely low-weight quantized neural networks with ReLU activation can approximate a class of specific functions well.

\begin{figure}[t]
\centering
\includegraphics[width=0.75\linewidth]{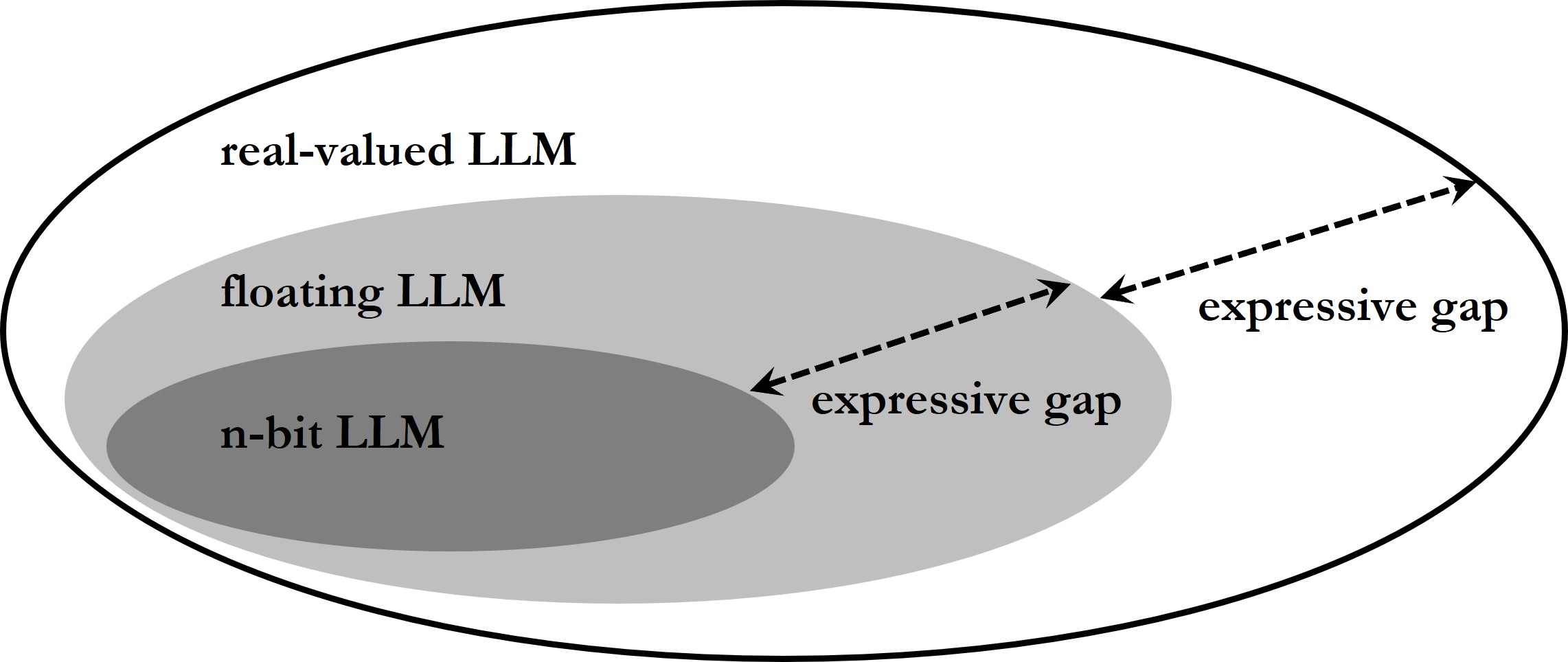}
\caption{Illustrations of expressive power and the corresponding expressive gap of real-valued, floating, and $n$-bit LLMs.}
\label{fig:expressive}
\end{figure}

Another important characterization is the expressive degradation of a weight-quantized model led by the decreasing number of quantization bits. Intuitively, the lower-bit weight quantization may lead to lower model performance regarding low-precision quantized weights as an approximation of real-valued ones~\citep{ding2019universal,yang2020:search,chen2024:ternary}. As shown in Figure~\ref{fig:expressive}, there may exist expressive gaps between real-valued, floating, and weight-quantized models, where the floating format indicates the commonly used number of bits, like 16 or 32 bits, while the $n$-bit format refers to lower-precision operations. In contrast, some researchers argue that weight quantization causes expressive degradation~\citep{ouyang2024:wq,ma2025:wq}. One argument regards the low-bit operation of weight quantization as the model noise; perhaps random noise, as a regularizer, may not harm the model performance~\citep{chatterjee2017:search,guo2018survey}.

In this paper, we conduct a theoretical investigation into the expressive capacity of LLMs relative to the number of quantization bits. More specifically, we aim to assess the strengths and limitations of weight quantization in LLMs by addressing the following three fundamental questions
\begin{enumerate}
	\item Can LLMs with quantized weights still achieve universal approximation?
	\item Whether and to what extent does the expressive power of a weight-quantized LLM degrade as the number of quantization bits decreases?
	\item What is the limiting precision of weight quantization of LLMs, as practical advances steadily drive weight quantization to increasingly lower bits?
\end{enumerate}

To tackle these three questions, we take an in-depth analysis on the expressive capacity of attention and Multi-Layer Perceptron (MLP) modules, which are two key components of LLMs. We firstly investigate the universal approximation and architecture complexity of weight-quantized LLMs with $n$-bit weights and present the following conclusions.

\begin{theorem}[Universal Approximation] \label{thm:UA_informal}
	Let $f_{\textrm{$n$-bit}}: \mathbb{R}^N \to \mathbb{R}^M$ be the function expressed by a weight-quantized LLM that consists of the attention and MLP modules in sequence, equipped with the ReLU activation and $n$-bit weights, where $n \in \mathbb{N}^+$. Let $H\in \mathbb{N}^+$ be an intermediate index that denotes both the output dimension of the attention module and the input dimension of the MLP module. The following results hold  
\begin{itemize}
    \item[i)] For $n > 1$, the collection of functions expressed by a $n$-bit weight-quantized LLM that consists of a single-head attention with the Key dimension of $\mathcal{O}(H)$ and Value dimension of $\mathcal{O}(H)$ and an MLP with a width of at most $H+M+\mathcal{O}(\|\boldsymbol{x}\|_{\infty})$ is dense in $\mathcal{C}(K \subseteq \mathbb{R}^N, \mathbb{R}^M)$ with respect to the uniform norm, where $\|\boldsymbol{x}\|_{\infty}$ denotes the infinity norm of an input variable $\boldsymbol{x}$.
    \item[ii)] For $n=1$, there exist a function $f: [-1,1]^N \to \mathbb{R}^M$ and a certain constant $\delta$, such that
    \[
    \sup_{\boldsymbol{x} \in [-1,1]^N} \| f(\boldsymbol{x}) - f_{\textrm{1-bit}}(\boldsymbol{x}) \| \geq \delta \ ,
    \]
for any 1-bit weight-quantized LLM $f_{\textrm{1-bit}}$ that each connection weight belongs to $\{0,1\}$.
\end{itemize}
\end{theorem}
Theorem~\ref{thm:UA_informal} examines the universal approximation of $n$-bit weight-quantized LLMs, in which one equipped with a narrow and considerably deep architecture maintains the universal approximation property for $n> 1$. This result provides positive responses to the first fundamental question. Table~\ref{tab:UA_summary} lists several remarkable achievements and our results on universal approximation and architecture complexity of the attention and MLP modules.

Besides, Theorem~\ref{thm:UA_informal} also shows that there exist certain counterexample functions living on $[-1,1]^N$ that cannot be approximated well by 1-bit weight-quantized LLMs even with exponential width and depth. This result derives an insight that the ternary format, or equally 1.58-bit where $1.58=\text{log}_2 3$, is the limiting precision for weight quantization. Notice that this argument also coincides with the practical advances of Microsoft research teams~\citep{wu2025bitnet,zhang2025:ternaryclip}. Therefore, we take responses to the third question.

\begin{table}[!htb]
\centering
\caption{Studies on universal approximation and architecture complexity of LLMs.}
\label{tab:UA_summary}
\begin{tabular}{l | l | c|c | l}
\toprule[1.5pt]
\textbf{Studies}  & \makecell{\textbf{Models}\\ \textbf{Activation}}  &  \makecell{\textbf{Weight}\\ \textbf{Formats}}  &\makecell{\textbf{Universal}\\ \textbf{Approximation}}  & \makecell{\textbf{Architecture}\\ \textbf{Complexity}} \\  \midrule[1pt]
\citet{hornik1989:UA}
& $\begin{aligned}
	&\text{MLP} \\
    &\text{non-polynomial}
\end{aligned}$ & \text{real-valued} & $\checkmark$ & $\begin{aligned}
    &\text{exponential width,} \\
    &\text{finite depth}
\end{aligned}$ \\  \midrule
\citet{gonon2023:UA}
& $\begin{aligned}
	&\text{MLP} \\
    &\text{ReLU}
\end{aligned}$ & \text{floating} & $\checkmark$ & $\begin{aligned}
    &\text{finite width} \\
    &\text{exponential depth}
\end{aligned}$ \\  \midrule
Our work (Theorem~\ref{thm:UA_mlp})
& $\begin{aligned}
	&\text{MLP} \\
    &\text{ReLU}
\end{aligned}$ & \text{$n$-bit} & $\checkmark$ & $\begin{aligned}
    &\text{linear width} \\
    &\text{exponential depth}
\end{aligned}$ \\  \midrule[1pt]
\citet{cheng2025:transformer}
& $\begin{aligned}
	&\text{Transformer} \\
	&\text{Softmax}
\end{aligned}$ & \text{real-valued} & $\checkmark$ & $\begin{aligned}
	&\text{polynomial width} \\
	&\text{finite depth}
\end{aligned}$ \\   \midrule
Our work (Theorem~\ref{thm:UA_att}) 
& $\begin{aligned}
	&\text{Attention} \\
	&\text{ReLU}
\end{aligned}$ & \text{$n$-bit} & $\checkmark$ & $\begin{aligned}
	&\text{linear width} \\
	&\text{exponential depth}
\end{aligned}$ \\
\bottomrule[1.5pt]
\end{tabular} 
\end{table}

\begin{theorem}[Expressive Degradation] \label{thm:UA_bound_informal}
Let $f_{Q_n(\theta)}( \boldsymbol{x} )$ and $f_\theta( \boldsymbol{x} )$ separately denote the functions expressed by LLMs with $n$-bit and real-valued weights, where $\boldsymbol{x} \in [-D,D]^N$ for $D>0$ and $Q_n(\theta)$ indicates the $n$-bit weight variable that corresponds to the real-valued $\theta$. Provided a $N_h$-head attention block and an $L$-layer MLP with the ReLU activation, there exist
\[
\epsilon>0
\quad\text{and}\quad
\delta = \mathcal{O}\left( H^{-1} N_h^{-1} n^{-(L+3)} \epsilon \right) \ ,
\]
such that if $\max_\theta | Q_n(\theta) - \theta | \leq \delta$, then the following holds
\[
\max_{\boldsymbol{x}} \max_\theta~ \| f_{Q_n(\theta)}( \boldsymbol{x} ) - f_\theta( \boldsymbol{x} ) \|_2 \leq \epsilon \ .
\]
\end{theorem}
Theorem~\ref{thm:UA_bound_informal} affirms the expressive degradation induced by the weight quantization and further shows that the approximation gap between weight-quantized and real-valued LLMs would decrease polynomially as the number of quantization bits increases. This theorem answers the second question, the bound of which reveals the expressive effect led by the number of quantization bits. 

The rest of this paper is organized as follows. Section~\ref{sec:pre} shows the useful terminologies and related studies of weight quantization. Section~\ref{sec:proof_sketch} highlights the proof sketches of our main conclusions. Section~\ref{sec:main}  completes theoretical theorems and the corresponding proofs. Section~\ref{sec:experiments} conducts numerical experiments. Section~\ref{sec:conclusions} concludes this work with discussions and prospects.

\section{Preliminaries}  \label{sec:pre}
This subsection consists of the useful notations in Subsection~\ref{subsec:notations}, the formulation of weight quantization in Subsection~\ref{subsec:wq}, and related studies in Subsection~\ref{subsec:rw}.

\subsection{Notations}  \label{subsec:notations}
Let $[N] = \{1, 2, \dots, N\}$ be an integer set for $N \in \mathbb{N}^+$, and $|\cdot|_{\#}$ denotes the number of elements in a collection, e.g., $|[N]|_{\#} = N$. We denote the preceding operation by the symbol $\preccurlyeq$, in which $\boldsymbol{x} \preccurlyeq 0$ given $\boldsymbol{x} \in \mathbb{R}^n$ means that every element $x_i \leq 0$ for any $i \in [n]$. 

\noindent\textbf{Vector and Matrix Norms.} We also consider the two typical norms of vectors or matrices. For $\mathbf{W} \in \mathbb{R}^{n \times m}$, we denote by
\[
\| \mathbf{W} \|_2 = \left( \sum_{i=1}^n \sum_{j=1}^m |\mathbf{W}_{ij}|^2 \right)^{1/2}
\]
and $\| \mathbf{W} \|_\infty = \max_{i,j} |\mathbf{W}_{ij}|$. Here, we only introduce the norms of $\| \cdot \|_2$ and $\| \cdot \|_\infty$. It is evident that the 2-norm can be bounded by the infinity one, i.e., $ \| \boldsymbol{w} \|_2 \leq \sqrt{n} \| \boldsymbol{w} \|_\infty$.

\noindent\textbf{Functional Space.} This work describes the expressive power of deep learning models in the Sobolev space within functional norms. Let $f_i$ be a scalar function from $K \subseteq \mathbb{R}^n$ to $\mathbb{R}$. Given $\boldsymbol{\alpha} = (\alpha_1, \alpha_2, \dots, \alpha_l)^{\top} \in \mathbb{N}^m$ and $\boldsymbol{x} = (x_1, x_2, \dots, x_n) \in K$, we define 
\[ 
D^{\boldsymbol{\alpha}} f_i(\boldsymbol{x}) = \frac{\partial^{\alpha_1}}{\partial x^{\alpha_1}} \frac{\partial^{\alpha_2}}{\partial x^{\alpha_2}} \dots \frac{\partial^{\alpha_l}}{\partial x^{\alpha_l}} f_i(\boldsymbol{x}) \ .
\]
We define the space of continuous functions $\mathcal{C}^q(K, \mathbb{R})$ for $q \in \mathbb{N}^+$ by a collection of $f_i$, where $f_i \in \mathcal{C}(K, \mathbb{R})$ and $D^r f_i \in \mathcal{C}(K, \mathbb{R})$ for $r \in [q]$. Let $\mu$ be a Lebesgue measure defined on $K$. Further, we define the Lebesgue spaces for the mapping $f: K \to \mathbb{R}^m$, in which $\mathcal{L}_\mu^p(K,\mathbb{R}^m)$ for $1\leq p < \infty$ and $\mathcal{L}_{\mu}^\infty(K,\mathbb{R}^m)$ for $p=\infty$, where $f \in \mathcal{C}(K,\mathbb{R}^m)$ and
\[
\left\| f \right\|_{L_\mu^p(K,\mathbb{R}^m)} \overset{\mathrm{def}}{=} \left(\int_{K} \|f(\boldsymbol{x})\|_2^p \dif \mu(\boldsymbol{x}) \right)^{1/p} < \infty
\quad\text{or}\quad
\left\| f \right\|_{L_\mu^\infty(K,\mathbb{R}^m)} \overset{\mathrm{def}}{=} \esssup_{\boldsymbol{x} \in K} \|f(\boldsymbol{x})\|_{\infty} < \infty \ .
\]
It is evident that $ \left\| f \right\|_{L_\mu^p(K,\mathbb{R}^m)} \leq \sqrt{m ~\mu(K)} \left\| f \right\|_{L_\mu^\infty(K,\mathbb{R}^m)}$. In general, we denote the Sobolev space by $\mathcal{W}^{q,p}_{\mu}(K,\mathbb{R}^m)$, defined as the collection of all functions $f \in \mathcal{C}^q(K, \mathbb{R}^m)$ and $D^{\boldsymbol{\alpha}} f \in \mathcal{L}_\mu^p(K,\mathbb{R}^m)$ for all $|\boldsymbol{\alpha}| \in [q]$.

\noindent\textbf{Algorithmic Complexity.} Given two functions $g,h\colon \mathbb{N}^+\rightarrow \mathbb{R}$, we denote by $h=\Theta(g)$ if there exist positive constants $c_1,c_2$, and $n_0$ such that $c_1g(n) \leq h(n) \leq c_2g(n)$ for every $n \geq n_0$; $h=\mathcal{O}(g)$ if there exist positive constants $c$ and $n_0$ such that $h(n) \leq cg(n)$ for every $n \geq n_0$; $h=\Omega(g)$ if there exist positive constants $c$ and $n_0$ such that $h(n) \geq cg(n)$ for every $n \geq n_0$.

\subsection{Weight Quantization of Transformer}   \label{subsec:wq}
In this subsection, we formulate a Transformer that constructs the basic block of LLMs as follows
\begin{equation} \label{eq:transformer}
\boldsymbol{y} = \text{MLP}(\boldsymbol{z}) 
\quad\text{and}\quad
\boldsymbol{z} = \text{ATT}(\boldsymbol{x}) \ ,
\end{equation}
where $(\boldsymbol{x},\boldsymbol{z},\boldsymbol{y}) $ indicates a pair of (input, hidden, output) variables. It is observed that the Transformer comprises the attention and MLP modules, corresponding to the \text{ATT}($\cdot$) and \text{MLP}($\cdot$) in Eq.~\eqref{eq:transformer}, respectively. Here, we omit the layer normalization for simplicity and generalize the formulations of the ATT and MLP modules. The first one is the ATT($\cdot$) module as follows
\begin{equation}  \label{eq:att}
\boldsymbol{z} = \mathbf{W}_\text{x}\ \boldsymbol{x} + \sum_{i=1}^{N_h} \mathbf{W}_\text{o}^{(i)} \left( \mathbf{W}_\text{v}^{(i)} \boldsymbol{x} + \boldsymbol{b}_\text{v}^{(i)} \right) ~\tau\left[ \left(\mathbf{W}_{\!\text{k}}^{(i)} \boldsymbol{x} + \boldsymbol{b}_\text{k}^{(i)} \right)^\top \left( \mathbf{W}_{\!\text{q}}^{(i)} \boldsymbol{x} + \boldsymbol{b}_\text{q}^{(i)} \right) \right] + \boldsymbol{b}_\text{o}^{(i)} \ ,
\end{equation}
where $(\boldsymbol{x},\boldsymbol{z}) \in ( \mathbb{R}^N , \mathbb{R}^H )$ is a pair of input and hidden variables, $\tau$ denotes the attention activation function, $N_h$ indicates the number of multi heads, and 
$
\mathbf{W}_\text{x} \in \mathbb{R}^{H \times N} ,
\mathbf{W}_\text{o}^{(i)} \in \mathbb{R}^{H \times N_v} ,
\mathbf{W}_\text{v}^{(i)} \in \mathbb{R}^{N_v \times N} ,
\mathbf{W}_{\!\text{k}}^{(i)} \in \mathbb{R}^{N_k \times N} ,
\mathbf{W}_{\!\text{q}}^{(i)} \in \mathbb{R}^{N_k \times N} ,
\boldsymbol{b}_\text{v}^{(i)} \in \mathbb{R}^{N_v} , 
\boldsymbol{b}_\text{k}^{(i)} \in \mathbb{R}^{N_k} ,
\boldsymbol{b}_\text{q}^{(i)} \in \mathbb{R}^{N_k} ,
\boldsymbol{b}_\text{o}^{(i)} \in \mathbb{R}^{H} ,
$
in which $N_v$ and $N_k$ are two hyper-parameters. Note that Eq.~\eqref{eq:att} adapts a generalized attention formulation that employs the bias terms $(\boldsymbol{b}_\text{v}^{(i)} , \boldsymbol{b}_\text{k}^{(i)} , \boldsymbol{b}_\text{q}^{(i)} , \boldsymbol{b}_\text{o}^{(i)} )$,  the linear concatenation $\mathbf{W}_\text{o}^{(i)}$ that follows \citet{yun2019:transformers}, and the dimension-scaling parameters $\mathbf{W}_\text{x}$. It is evident that Eq.~\eqref{eq:att} can be degenerated to the original one~\citep{vaswani2017:attention} if the bias and concatenation terms are set as 1. The second one is the MLP($\cdot$) module, equipped with an $L$-layer architecture where $L \in \mathbb{N}^+$
\begin{equation}  \label{eq:mlp}
\left\{~\begin{aligned}
&\boldsymbol{h}^{(0)} = \boldsymbol{z} \ , \\
&\boldsymbol{h}^{(l)} = \sigma\left( \mathbf{W}_\text{mlp}^{(l)} \boldsymbol{h}^{(l-1)} + \boldsymbol{b}_\text{mlp}^{(l)} \right)  \quad\text{for}\quad l \in [L] \ , \\
& \boldsymbol{y} = \mathbf{W}_\text{z}\ \boldsymbol{z} + \boldsymbol{h}^{(L)} \ ,  
\end{aligned}\right.
\end{equation}
where $(\boldsymbol{z},\boldsymbol{y}) \in ( \mathbb{R}^H , \mathbb{R}^M )$ is a pair of hidden and output variables, $\sigma$ is the MLP activation function, and  $\mathbf{W}_\text{z}$ denotes the dimension-scaling parameter. Notice that we here employ the superscripts $(i)$ and $(l)$ to record the multi-head-wise and layer-wise indices, respectively. 

In this work, all connection terms $(\mathbf{W}_\text{x},\mathbf{W}_\text{z}, \mathbf{W}_\text{o}^{(i)},
\mathbf{W}_\text{v}^{(i)},
\mathbf{W}_{\!\text{k}}^{(i)},
\mathbf{W}_{\!\text{q}}^{(i)}, \mathbf{W}_\text{mlp}^{(l)})$ and bias terms $(\boldsymbol{b}_\text{v}^{(i)} , \boldsymbol{b}_\text{k}^{(i)} , \boldsymbol{b}_\text{q}^{(i)} , \boldsymbol{b}_\text{o}^{(i)}, \boldsymbol{b}_\text{mlp}^{(l)})$ are named as learnable weights and are ideally set to obey the real-valued field. In practice, developers often format learnable weights as a floating number. Correspondingly, we use $\theta$ to indicate the learnable weight variable and denote the floating operation by $F$ so that $F(\theta)$ falls in the floating format. We further consider the weight quantization, i.e., the learnable weights are quantized into the $n$-bit format $(\mathcal{P}_n,\mathcal{N}_n)$ in which $n$ is the number of quantization bits, $\mathcal{P}_n$ indicates the collection of $n$-bit positions, and $\mathcal{N}_n$ denotes the quantized values correspondingly. In detail, we have $\mathcal{P}_1 = \{ 0 , 1 \} $ and $\mathcal{N}_1 = \{0,1\}$ for the case of $n=1$; $\mathcal{P}_n = \{ 0 , 1 , \dots, n-1, \omega \}$ and $\mathcal{N}_n = \{0,\pm1,\dots,\pm (n-1)\}$ for the case of $n\geq 2$, in which $\omega$ is the inverse operation of addition. For convenience, we employ the symbol $Q_n$ to denote the $n$-bit quantization operation, so that $Q_n: \mathbb{R} \to \mathcal{N}$ where $Q_n(\theta) \in \mathcal{N}_n$. We also ignore the codebook storage due to the lower order of complexity and only consider the ReLU activation. In general, we consider real-valued inputs, but their storage format should be compatible with bit-wise operations. Throughout this paper, we denote the functions expressed by neural networks with real-valued, floating, and $n$-bit weights by $f_{\theta}$, $f_{F(\theta)}$, and $f_{Q_n(\theta)}$, respectively. The floating format corresponds to 32-bit in the experiments.

\subsection{Related Works} \label{subsec:rw}
The weight quantization technique aims to encode model weights into a format with fewer bits, where extremely low-bit operations typically result in significant model compression and inference acceleration. Thus, in recent years, weight quantization has obtained increasing interest in the fields of lightweight computations and resource-constrained applications. BinaryConnect~\citep{cour2015:optimization} removes about 2/3 of the multiplications, thus empirically resulting in a 3x speed-up in training time and at least 16x memory savings. Ternary Weight Networks~\citep{li2016ternary} achieves up to a 16× model compression rate. To minimize the accuracy degradation led by weight ternarization, Incremental Network Quantization~\citep{zhou2017incremental} introduces a new computation mechanism that consists of weight partition, group-wise quantization, and retraining. In contrast, \citet{yang2020:search} utilized a differential method to search ternary weights. \citet{chatterjee2017:search} explored the optimal choice of the number of quantization bits. \citet{ouyang2024:wq} and \citet{ma2025:wq} developed weight-quantized deep learning models with scaling laws. \citet{wu2025bitnet} presented a lightweight pipeline that fine-tunes off-the-shelf full-precision LLMs into ternary weights for specific downstream tasks. TernaryCLIP~\citep{zhang2025:ternaryclip} achieves extremely low-cost and high-efficiency computations in image-text contrastive modeling.

Despite the success of many weight-quantized models on real-world datasets, the theoretical understanding is still very limited. A theoretical characterization of weight quantization should answer questions about its approximation, optimization, and generalization. For approximation, \citet{yayla2021:ua} investigated the universal approximation of 1-bit neural networks with binary and real-valued inputs. \citet{ding2019universal} proved that extremely low-weight-quantized neural networks with ReLU activation can approximate a class of specific functions well and provided a theoretical complexity bound for estimating an optimal bit-width. For optimization, BinaryConnect employed STE to approximate the gradient, which is theoretically supported by approximate optimization~\citep{cour2015:optimization}. \citet{li2017:optimization} theoretically discussed the training method of weight-quantized neural networks, in which the training accuracy guarantee under the convexity assumption and the algorithm behavior in non-convex problems are analyzed. For generalization, \citet{stephan1997:vc} investigated the VC dimension of perceptrons with weights restricted to $\pm 1$. \citet{anderson2017:wq} analyzed the geometrical properties of 1-bit neural networks, in which data features are effectively captured in high-dimensional geometrical space. Some researchers argued that weight quantization works as a regularizer that contributes to generalization by regarding the low-bit operation of weight quantization as the noise of the primary model~\citep{chatterjee2017:search,guo2018survey}. \citet{gholami2021:generalization} emphasized the trade-off between quantization and model compression and generalization ability.

\section{Proof Sketches}  \label{sec:proof_sketch}

There is a common sense that the nonlinear computations and the scaling number of parameters promote the expressivity of deep learning models. Intuitively, the nonlinearity in MLPs originates from the nonlinear activation function with linear pre-activation, whereas the nonlinearity in Attention mechanisms arises from the product of a nonlinear function and the linear weights, where the former acts as a nonlinear activation function composed of the quadratic pre-activation, provided the following conversions
\begin{equation}   \label{eq:conversion}
	\left\{~\begin{aligned}
		&\text{ATT}:\quad \tilde{\boldsymbol{z}} \stackrel{\text{def}}{=} \boldsymbol{z} - \mathbf{W}_\text{x} \  \boldsymbol{x} = \sum_{i=1}^{N_h} \rho^{(i)} (\boldsymbol{x}) \left( \mathbf{W}_{\!\rho}^{(i)} \ \boldsymbol{x} + \boldsymbol{b}_\rho^{(i)} \right) + \boldsymbol{b}_\text{o}^{(i)}  \ , \\
		&\text{MLP}:\quad \tilde{\boldsymbol{y}} \stackrel{\text{def}}{=} \boldsymbol{y} - \mathbf{W}_\text{z} \ \boldsymbol{z} = \sigma\left( \mathbf{W}_\text{mlp}^{(L)} \boldsymbol{h}^{(L-1)} + \boldsymbol{b}_\text{mlp}^{(L)} \right)  \ ,  
	\end{aligned}\right.
\end{equation}
with
\[
\rho^{(i)} (\boldsymbol{x}) = \tau\left[ \left(\mathbf{W}_{\!\text{k}}^{(i)} \boldsymbol{x} + \boldsymbol{b}_\text{k}^{(i)} \right)^\top \left( \mathbf{W}_{\!\text{q}}^{(i)} \boldsymbol{x} + \boldsymbol{b}_\text{q}^{(i)} \right) \right] \ ,
\quad
\mathbf{W}_{\!\rho}^{(i)} = \mathbf{W}_\text{o}^{(i)} \mathbf{W}_\text{v}^{(i)} \ ,  
\quad
\boldsymbol{b}_\rho^{(i)} = \mathbf{W}_\text{o}^{(i)} \boldsymbol{b}_\text{v}^{(i)} \ .
\]
Intuitively, the ATT module looks like a two-layer MLP, where the first layer takes an identity function as activation while the second layer employs the nonlinear function with quadratic pre-activation as the connection weight. Thus, it is reasonable to deduce that ATT and MLP are functionally analogous from the perspective of expressive capacity and share common characteristics between the two architectures. Therefore, we focus more on the effects of weight quantization on the activated units with linear and quadratic pre-activations, that is, $\rho^{(i)} (\cdot)$ and $\sigma(\cdot)$ in Eq.~\eqref{eq:conversion}. Specifically, we theoretically investigate the universal approximation and expressive degradation of weight-quantized ATT and MLP modules. 

\begin{table}[t]
	\caption{The useful operations (Ops.) and corresponding programming.}
	\centering
	\label{tab:bits}
	\begin{tabular}{ll|ll}
		\toprule
		Ops. & Basic Programming  & Ops. & Composed Programming   \\  \midrule
		$\ll p$  & shift right by $p$ bits 
		& $\land$   &   XOR \quad $a \land b = (a | b) \& (\sim a | \sim b)$ \\
		$\gg q$  & shift left by $q$ bits  
		& $+$         &   Addition \quad  $a + b = (a\land b) | ((a \& b) \ll 1) $ \\
		$\&$    &  AND \quad  e.g., $0 \& 1 = 0$ 
		&~$\cdot$   &   Multiplication \quad  $1 \cdot a = a $ \\
		$|$       &  OR \quad  e.g., $0 | 1 = 1$  \\
		$\sim$ &  NOT \quad  e.g., $\sim 1 = 0$ & &  \\
		\bottomrule                     
	\end{tabular} 
\end{table}

\begin{figure*}[t]
	\centering
	\includegraphics[width=1\linewidth]{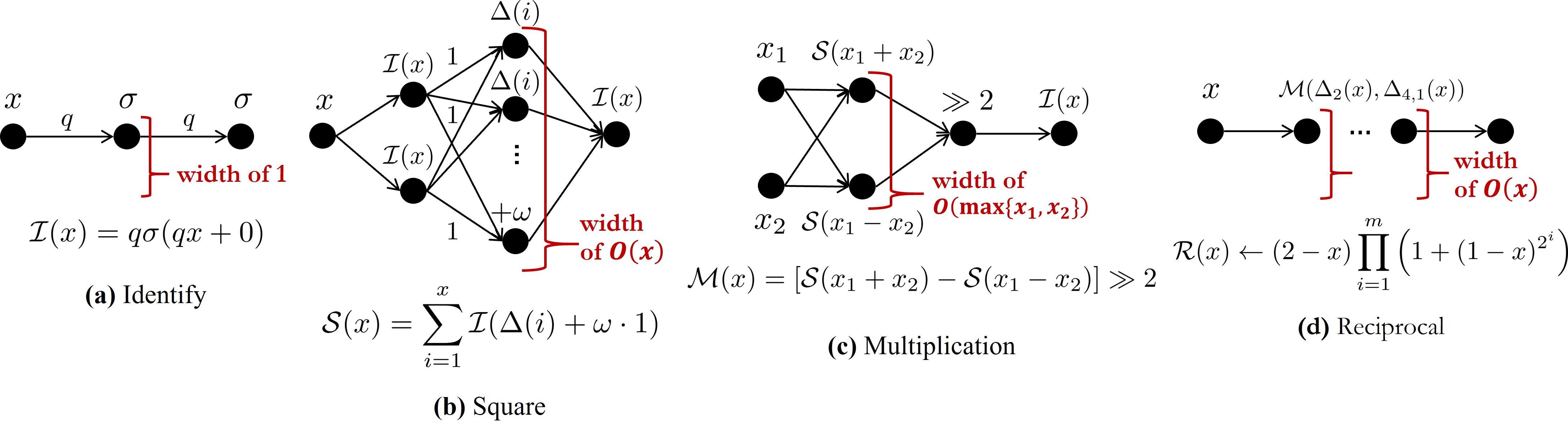}
	\caption{A schematic diagram of LLM architectures that approximate four hierarchical functions.}
	\label{fig:hierachical}
\end{figure*}

Our primary objective is to approximate the universal family of target functions through a concerned module with quantized weights. In conventional approximation theories~\citep{hornik1989:UA,yun2019:transformers,gonon2023:UA}, the universal family of target functions usually adapts the space of continuous functions from $\mathbb{R}^N$ to $\mathbb{R}^H$ or from $\mathbb{R}^H$ to $\mathbb{R}^M$, that is, $\mathcal{C}(K_{\text{in}},\mathbb{R}^H)$ where $K_{\text{in}} \subseteq \mathbb{R}^N$ or $\mathcal{C}(K_\text{h},\mathbb{R}^M)$ where $K_\text{h} \subseteq \mathbb{R}^H$. Accordingly, the vector-valued functions approximated by the ATT or MLP modules can be equivalently decomposed as the corresponding scalar-valued ones. Thus, our objective becomes to approximate $\mathcal{C}(K_{\text{in}},\mathbb{R})$ and $\mathcal{C}(K_\text{h},\mathbb{R})$ using the functions expressed by the ATT and MLP modules with quantized weights, respectively. We corroborate this objective by deconstructing the universal family of target scalar-valued functions as a collection of sub-functions via Taylor expansion, i.e., 
\begin{equation}  \label{eq:taylor}
	f(x) = \sum_{n=0}^{\infty} \frac{f^{(n)}(0)}{n!} x^n = f(0) + f'(0)x + \frac{f''(0)}{2!}x^2 + \dots + \frac{f^{(n)}(0)}{n!}x^n + R_n(x) \ .
\end{equation}
Based on the Drawer principle~\citep{kidger2020:ua}, the sub-functions are with polynomial formations and contingent upon only four hierarchical functions, involving
\[
\left\{ ~\begin{aligned}
\text{identify}: &\quad \mathcal{I}(x) = x \ , \\
\text{square}: &\quad \mathcal{S}(x) = x^2 \ , \\
\text{multiplication}: &\quad \mathcal{M}(x_1,x_2) = x_1x_2 \ , \\
\text{reciprocal}: &\quad \mathcal{R}(x) = 1/x \ , \\
\end{aligned} \right.
\]
where the square and multiplication functions implicate the operation that upgrades the power index of polynomial functions, while the reciprocal function leads to the operation that degrades the power index of polynomial functions. Thus, the primary issue of approximating the universal family of target functions through the weight-quantized MLP and ATT modules can be converted into another problem that approximates the above four hierarchical functions by the activated units with linear and quadratic pre-activations with quantized weights. To solve this problem, we program the hierarchical functions and weight-quantized activations through bitwise operations acting on bit strings.  Table~\ref{tab:bits} lists the useful bitwise operations, and Figure~\ref{fig:hierachical} illustrates the schematic diagram of LLM architectures that approximate four hierarchical functions.

It suffices to program the activation function through bitwise operations, since the linear and quadratic pre-activations comprise multiplication and addition operations that can be easily implemented. For the activated computation $\boldsymbol{h} = \text{ReLU}( \boldsymbol{s} )$ with the ReLU function, it is intuitive to exploit the disjunctive program with bitwise operations. Hence, the ReLU activation equals to
\begin{equation*}
	\begin{bmatrix}
		\boldsymbol{h} = 0 \\
		\boldsymbol{s} \preccurlyeq 0 
	\end{bmatrix}  \bigvee
	\begin{bmatrix}
		\boldsymbol{h} = \boldsymbol{s} \\
		\boldsymbol{s} \succcurlyeq 0 
	\end{bmatrix} \ ,
\end{equation*}
where 
\[
\boldsymbol{s} = \left(\mathbf{W}_{\!\text{k}}^{(i)} \boldsymbol{x} + \boldsymbol{b}_\text{k}^{(i)} \right)^\top \left( \mathbf{W}_{\!\text{q}}^{(i)} \boldsymbol{x} + \boldsymbol{b}_\text{q}^{(i)} \right)
\quad\text{for}\quad i \in [N_h]
\]
and
\[
\boldsymbol{s} = \mathbf{W}_\text{mlp}^{(l)} \boldsymbol{h}^{(l-1)} + \boldsymbol{b}_\text{mlp}^{(l)}  \quad\text{for}\quad l \in [L] \ .
\]
This disjunctive program presents the auxiliary variables for each disjunction and can be implemented according to \citep{aftabi2024feed,hunt2014artificial}. Here, we provide an alternative approach by formulating the ReLU activation as a set of linear inequalities. The ReLU activation can be formulated as follows
\begin{equation}  \label{eq:inequality}
	\left\{~\begin{aligned}
		& \text{Sampling initial variables from }  [B_l,B_u]^{N} \ , \\
		& \boldsymbol{h}  \succcurlyeq  0 \ , \\
		& \boldsymbol{h}  \succcurlyeq  \boldsymbol{s} \ , \\
		& \boldsymbol{h}  \preccurlyeq  \boldsymbol{s} + \left( \boldsymbol{1} - \boldsymbol{q} \right) \odot \boldsymbol{m}\ , \\
		& \boldsymbol{h}  \preccurlyeq  \boldsymbol{q} \odot \boldsymbol{m} \ , 
	\end{aligned}\right.
\end{equation}
where $B_l$ and $B_u$ separately are the lower and upper bounds of input variables, $\boldsymbol{m}$ indicates the programming vector, $\boldsymbol{q} \in \{0,1\}^{n_l}$, and $\odot$ denotes the element-wise multiplication. Notice that there have been several mature linear programming algorithms that can solve Eq.~\eqref{eq:inequality}. The efficiency of solving these inequalities is significantly affected by the relaxation of linear programming, which is determined by the programming vector $\boldsymbol{m}$ since it impacts how tightly the linear programming relaxation estimates the convex hull of the feasible region~\citep{aftabi2024feed}. \citet{bunel2018unified} proposed a specific solution procedure by establishing the initial bounds $B_l$ and $B_u$ of MLPs and propagating the bounds layer by layer through a process known as interval arithmetic. Therefore, the weight-quantized LLMs can be implemented by integer programs and bitwise operations.

\begin{algorithm}[!htb]
	\caption{Compute $\mathcal{S}(x)$ and $\mathcal{R}(x)$ using Bit-wise Operations.}
	\label{alg:bit_wise_fcn}
	\begin{algorithmic}[1]
		\STATE \textbf{Input:} $x$
		\STATE \textbf{Output:} $\mathcal{S}(x)$ and $\mathcal{R}(x)$
		\IF{$x == 0$}
		\STATE\textbf{return} 0
		\ENDIF
		\IF{$x < 0$}
		\STATE $x \gets -x$
		\ENDIF
		\STATE $\mathcal{S}(x) \gets 0$
		\STATE $p \gets 0$ 
		\WHILE{$x > 0$}
		\IF{$x \& 1$}
		\STATE $\mathcal{S}(x) \gets \mathcal{S}(x) + (x \ll p)$
		\ENDIF
		\STATE $x \gets x \gg 1$
		\STATE $p \gets p + 1$
		\ENDWHILE
		\STATE\textbf{RETURN} $\mathcal{S}(x)$
		\STATE $\Delta_1(x) \gets \mathcal{I} (\omega x + 1)$
		\STATE $\Delta_2(x) \gets 1 \cdot \Delta_1(x) + 1$
		\STATE $\Delta_3(x) \gets \mathcal{S}( \Delta_1(x) ) $
		\STATE $\Delta_{4,1}(x) \gets \mathcal{I} \left( 1 \cdot \mathcal{S} (\Delta_3(x)) + 1  \right)$
		\STATE $\mathcal{R}(x) \gets \mathcal{M} \left(   \Delta_2(x)  ,  \Delta_{4,1}(x)  \right) $
		\FOR{$i = 1:m-1$}
		\STATE $\Delta_{4,i+1}(x) \gets \mathcal{I} \left( 1 \cdot \mathcal{S}^{i+1} (\Delta_3(x)) + 1  \right) $
		\STATE $\mathcal{R}(x) \gets \mathcal{M} \left(   \mathcal{R}(x)  ,  \Delta_{4,i+1}(x)  \right) $
		\ENDFOR
		\STATE\textbf{RETURN} $\mathcal{R}(x)$
	\end{algorithmic}
\end{algorithm}

\clearpage
\section{Main Results}  \label{sec:main}
In this section, we formally propose our main results, including the universal approximation of $n$-bit ATTs and MLPs for $n \geq 2$ in Subsection~\ref{subsec:proof_01}, the expressive collapse of 1-bit ATTs and MLPs in Subsection~\ref{subsec:proof_02}, and the expressive degradation of weight-quantized ATTs and MLPs led by a decreasing number of quantization bits in Subsection~\ref{subsec:proof_03}.

\subsection{Universal Approximation}   \label{subsec:proof_01}
Now, we present our first theorem as follows.
\begin{theorem} \label{thm:UA_mlp}
Let $K_\text{h} \subseteq \mathbb{R}^H$ and $n\geq 2$. Provided the ReLU activation, the collection of functions expressed by a $n$-bit weight-quantized MLP with a width of at most $H+M+\mathcal{O}(\|\boldsymbol{x}\|_{\infty})$ is dense in $\mathcal{C}(K_\text{h} \subseteq \mathbb{R}^H, \mathbb{R}^M)$ with respect to the uniform norm, where $\|\boldsymbol{x}\|_{\infty}$ denotes the infinity norm of an input variable $\boldsymbol{x} \in K_\text{h}$.
\end{theorem}
Theorem~\ref{thm:UA_mlp} shows the universal approximation properties of the weight-quantized MLPs equipped with a narrow and considerably deep architecture. This result is a qualitative difference compared to those of shallow architectures in Table~\ref{tab:UA_summary}. 

The proof of Theorem~\ref{thm:UA_mlp} is based on the approximation of MLPs to four hierarchical functions and the Drawer principle. Any continuous function can be approximated well by a polynomial function from Eq.~\eqref{eq:taylor}. Thus, it suffices to prove that a deep MLP composed of ReLU activation can represent any-order polynomial function. Following the decomposition idea of \citet{kidger2020:ua}, this work deconstructs the concerned polynomial function into four hierarchical functions, categorized as the identity, upgrade, and degrade operations. We further implement these hierarchical operations by exploiting several bit-wise operations mentioned in Table~\ref{tab:bits}. Along this line of thought, we provide several useful lemmas as follows, which inherit the conditions of Theorem~\ref{thm:UA_mlp}.

\begin{lemma} \label{lemma:hierarchical_mlp}
Let $f_{\textrm{1-bit}}(\cdot)$ denote the function expressed by a $n$-bit weight-quantized MLP. The following approximation properties hold
\begin{itemize}
	\item[(1)] $f_{\textrm{1-bit}}(\cdot)$ equipped with two layers and a width of 1 can exactly approximate the identity $\mathcal{I}: \mathbb{R} \to \mathbb{R}$ where $\mathcal{I}(x) = x$.
	\item[(2)] $f_{\textrm{1-bit}}(\cdot)$ equipped with two layers and a width of $\mathcal{O}(x)$ can exactly approximate the square $\mathcal{S}: \mathbb{R}  \to \mathbb{R}$ where $\mathcal{S}(x) = x^2$.
	\item[(3)] $f_{\textrm{1-bit}}(\cdot)$ equipped with one layer and a width of $\mathcal{O}(x)$ can exactly approximate the multiplication $\mathcal{M}: \mathbb{R} \times \mathbb{R} \to \mathbb{R}$ where $\mathcal{M}(x_1,x_2) = x_1 x_2$.
	\item[(4)] $f_{\textrm{1-bit}}(\cdot)$ equipped with one layer and a width of 2 can uniformly approximate the reciprocal $\mathcal{R}: \mathbb{R} \to \mathbb{R}$ where $\mathcal{R}(x) = 1/x$.
\end{itemize}
\end{lemma}
Lemma~\ref{lemma:hierarchical_mlp} shows the width complexities of $n$-bit weight-quantized MLPs that approximate several hierarchical operations within $\epsilon>0$ on the compact set $K_\text{h}$. The full proof of Lemma~\ref{lemma:hierarchical_mlp} can be accessed from Appendix~\ref{app:lemma_hierarchical_mlp}.

\begin{figure*}[!htb]
	\centering
	\includegraphics[width=0.8\linewidth]{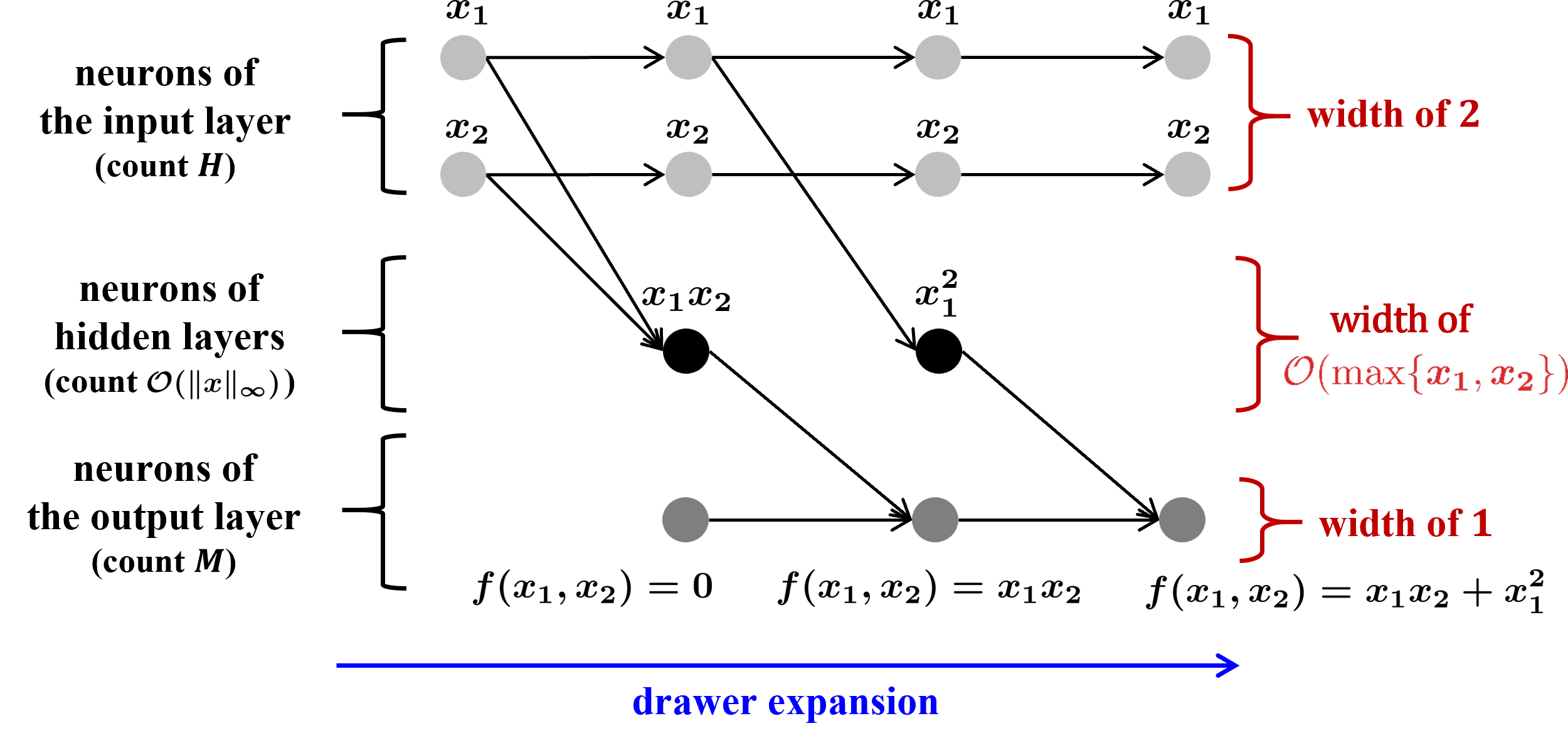}
	\caption{The example of approximating  $f(x_1,x_2) = x_1 x_2 + x_1^2$ for illustrating the width computing of weight-quantized MLPs based on the Drawer principle. }
	\label{fig:ua}
\end{figure*}

Based on the above results, we can establish an apposite neural network in which (1) only one hierarchical operation is performed on each hidden layer using at most $\mathcal{O}(\|\boldsymbol{x}\|_{\infty})$ hidden neurons and (2) each input or output is recorded by one hidden neuron. Figure~\ref{fig:ua} illustrates the width computing procedure through an example of approximating $f(x_1,x_2) = x_1 x_2 + x_1^2$, where each neuron of ``inputs'' conserves the original input layer using the identity operation, each neuron of the output layer records the computing result, and the neuron of hidden layers derives the upgrade or degrade calculation. According to the Drawer principle, the width equals the sum of the numbers of inputs, outputs, and positions, that is, $H+M+\mathcal{O}(\|\boldsymbol{x}\|_{\infty})$. In this figure, the neurons of the input layer conserve the original inputs $x_1$ and $x_2$. The first hidden layer computes the multiplication function $x_1 x_2$ using the width of $\mathcal{O}(\max\{x_1, x_2\})$, while the second hidden layer calculates the square function $x_1^2$ using the width of $\mathcal{O}(x_1)$. The neurons of the output layer record the computing results from the hidden layers. Consequently, $f(x_1,x_2)$ can be well approximated by a $n$-bit weight-quantized MLP with a width of $2+1+\mathcal{O}(\max\{x_1, x_2\})$, coinciding with the result of Theorem~\ref{thm:UA_mlp}. The full proof of Theorem~\ref{thm:UA_mlp} can be obtained in Appendix~\ref{app:thm_UA_mlp}.

\begin{theorem} \label{thm:UA_att}
	Let $K_\text{in} \subseteq \mathbb{R}^N$ and $n\geq 2$. Provided the ReLU activation, the collection of functions expressed by a $n$-bit weight-quantized ATT with the configurations of $N_h = 1$, $N_v = \mathcal{O}(H) $, and $N_k = \mathcal{O}(N)$ is dense in $\mathcal{C}(K_\text{in}  \subseteq \mathbb{R}^N, \mathbb{R}^H)$ with respect to the uniform norm.
\end{theorem}
Theorem~\ref{thm:UA_att} shows that the weight-quantized ATTs with a linear architecture complexity maintain the universal approximation property. The proof idea of Theorem~\ref{thm:UA_att} is similar to that of Theorem~\ref{thm:UA_mlp}, necessitating the proof of the following lemma.
\begin{lemma} \label{lemma:hierarchical_att}
	If one equips the ATT module with the ReLU activation and $n$-bit weights where $n \geq 2$, then the identity, square, and multiplication functions can be approximated well by the one-layer ATT module with $N_h=1, N_v=1, N_k\leq 2$, while the reciprocal function can be uniformly approximated by the ATT module with $N_h=1, N_v=1, N_k=1$ and a considerably deep architecture. 
\end{lemma}
Lemma~\ref{lemma:hierarchical_att} shows the width complexities of $n$-bit weight-quantized ATTs for approximating several hierarchical functions. The full proof of Lemma~\ref{lemma:hierarchical_att} can be accessed from Appendix~\ref{app:lemma_hierarchical_att}.

\noindent\textbf{Remarks.} Since the Transformer is constructed by the ATT and MLP modules in sequence, the architecture complexity of the weight-quantized Transformer is the minimization of those of the weight-quantized ATT and MLP modules. By combining Theorem~\ref{thm:UA_mlp} with Theorem~\ref{thm:UA_att}, we can prove the first part of Theorem~\ref{thm:UA_informal}. $\hfill\square$

\subsection{Expressive Collapse of 1-Bit ATTs and MLPs}  \label{subsec:proof_02}
In this subsection, we investigate the expressive power of 1-bit ATTs and MLPs where each learnable weight entry belongs to $\{0,1\}$.
\begin{theorem} \label{thm:1-bit_mlp}
Let the function $f(\boldsymbol{x})$ map from $[-1,1]^N$ to $\mathbb{R}^H$ or from $[-1,1]^H$ to $\mathbb{R}^M$, which corresponds to the function $f_{\textrm{1-bit}}$ expressed by the weight-quantized ATT or MLP equipped with the ReLU activation and 1-bit weights, respectively. There exists a certain constant $\delta$, such that for any 1-bit weight,
 it holds $\sup_{\boldsymbol{x}} \| f(\boldsymbol{x}) - f_{\textrm{1-bit}}(\boldsymbol{x}) \| \geq \delta$.
\end{theorem}
Theorem~\ref{thm:1-bit_mlp} displays a negative result on the expressive power of ATTs and MLPs with 1-bit weights, in which there exists at least one function living on $[-1,1]^N$ or $[-1,1]^H$ that cannot be approximated well by 1-bit weight-quantized MLPs even with exponential width and depth.  

The key idea of proving Theorem~\ref{thm:1-bit_mlp} is to construct a counterexample for 1-bit weight-quantized ATTs and MLPs. Here, we take an example that corresponds to 1-bit MLPs. Let $x_i$ denote the $i$-th element of $\boldsymbol{x} \in [-1,1]^H$ for $i \in [H]$. For example, $x_1$ is the first element of $\boldsymbol{x}$. Here, we construct a counterexample function $f: [-1,1]^H \to \mathbb{R}$ as follows
\[
f(\boldsymbol{x}) = \left\{~ \begin{aligned}
	\exp\left(  \frac{cx_1^2}{ cx_1^2 +1 }  \right) \ , &\quad |x_1| \leq  0.5 \ , \\
	0 \qquad\quad , &\quad |x_1| > 0.5 \ ,
\end{aligned}
\right.
\]
where $c<0$. It is observed that $f$ is a non-negative function and maintains its maximum at the original point $\boldsymbol{x} = \boldsymbol{0}$. Provided the counterexample function, we intuitively observe that the maximum gap between the target function $f$ and its 1-bit approximation $f_{\textrm{1-bit}}$ over the input domain $[-1,1]^H$ is greater than that over $\{-1,0,1\}^H$, that is, $\max_{\boldsymbol{x} \in \{-1,0,1\}^H} | f(\boldsymbol{x}) - f_{\textrm{1-bit}}(\boldsymbol{x}) |$, where the optimization objective can be unfolded as 
\[
| f(\boldsymbol{x}) - f_{\textrm{1-bit}}(\boldsymbol{x}) | = \left\{~\begin{aligned}
| 1 - f_{\textrm{1-bit}}(\boldsymbol{x}) | ~, & \quad x_1 = 0 \ , \\
| 0 - f_{\textrm{1-bit}}(\boldsymbol{x}) | ~, & \quad x_1 \in \{-1,1\} \ .
\end{aligned} \right.
\]
It is observed that $f_{\textrm{1-bit}}(\boldsymbol{x}) \in \{-1,0,1\}$ for $\boldsymbol{x} \in \{-1,0,1\}^H$ and $f_{\textrm{1-bit}}(\boldsymbol{0})=0$, and thus $| f(\boldsymbol{x}) - f_{\textrm{1-bit}}(\boldsymbol{x}) | \in [0,2]$. Hence, we can measure the maximum gap by
\[
\begin{aligned}
\sup_{\boldsymbol{x} \in [-1,1]^H} \| f(\boldsymbol{x}) - f_{\textrm{1-bit}}(\boldsymbol{x}) \| 
&\geq \max_{\boldsymbol{x} \in \{-1,0,1\}^H} | f(\boldsymbol{x}) - f_{\textrm{1-bit}}(\boldsymbol{x}) | \\
&= \max_{x_2, \dots, x_H \in \{-1,0,1\}^{H-1}}  | 1 - f_{\textrm{1-bit}}(\boldsymbol{x}) | + | 0 - f_{\textrm{1-bit}}(\boldsymbol{x}) | \\
&\geq \max_{x_i \in \{-1,0,1\}^H}  | 1 - f_{\textrm{1-bit}}(\boldsymbol{x}) | + | f_{\textrm{1-bit}}(\boldsymbol{x}) | \\
&\geq 1 \ ,
\end{aligned}
\]
where $i \in \{2,\dots, H\}$. Finally, we make two remarks. First, the concerned counterexample function cannot be exactly approximated by $f_{\textrm{1-bit}}$ but can be approximated well by real-valued MLPs. Second, the results of Theorem~\ref{thm:1-bit_mlp} also hold in the case that the binary weight belongs to $\{-1,1\}$. The full proof can be obtained in Appendix~\ref{app:1-bit_mlp}.

\subsection{Expressive Degradation}   \label{subsec:proof_03}
This subsection investigates the expressive degradation led by the number of quantization bits. Now, we present the theoretical results relative to the expressive degradation of ATTs as follows.

\begin{theorem}  \label{thm:rate_att}
Let $K_\text{in}$ be a compact set in $[-D,D]^N$ where $D>0$, and $\mu$ is a probability measure defined on $K_\text{in}$, $f_{Q_n(\theta)}( \cdot )$ is a function expressed by an ATT with a number of multi heads at most $N_h$ and the ReLU activation. For any $\epsilon > 0$, if there exists 
\[
\delta_1 = \mathcal{O} \left( C_{\textrm{att}}^{-1} D^{-3} n^{-3} \epsilon \right)
\quad\text{with}\quad
C_{\textrm{att}} = H N^2 N_h N_k^{\frac{3}{2}} N_v^{\frac{3}{2}}
\]
satisfying $\max_\theta | Q_n(\theta) - \theta | \leq \delta_1$, it holds
\[
\left\|  f_{Q_n(\theta)}( \boldsymbol{x} ) - f_\theta( \boldsymbol{x} ) \right\|_{L_\mu^\infty(K_\text{in},\mathbb{R}^H)} \leq \epsilon \ .
\]
Further, if there exists $\delta_2 = \mathcal{O}( N^{3/2} \mu^3(K_\text{in}) ) \cdot \delta_1$ such that $\max_\theta | Q_n(\theta) - \theta | \leq \delta_2$, it holds
\[
\left\|  f_{Q_n(\theta)}( \boldsymbol{x} ) - f_\theta( \boldsymbol{x} ) \right\|_{L_\mu^2(K_\text{in},\mathbb{R}^H)} \leq \epsilon \ .
\]
\end{theorem}
Theorem~\ref{thm:rate_att} shows that the approximation gap between the weight-quantized and real-valued ATTs would decrease polynomially as the number of quantization bits increases, in which $n^{-3}$ reflects the expressive effect led by the number of quantization bits and $C_{\textrm{att}} = H N^2 N_h N_k^{{3}/{2}} N_v^{{3}/{2}}$ indicates the approximation effect led by architecture configurations. This conclusion holds in the Lebesgue space with the $L^\infty$ and $L^2$ norms. 

The proof idea of Theorem~\ref{thm:rate_att} is straightforward. It suffices to prove the Lipschitz continuity of an ATT module equipped with the ReLU activation with respect to a pair of adjacent parameters $\theta$ and $\hat{\theta}$, that is, the following holds
\[
\sup_{\boldsymbol{x} \in K_\text{in}} \left\| f_{\theta}( \boldsymbol{x} ) - f_{\hat{\theta}} (\boldsymbol{x}) \right\|_\infty \leq C_\theta \ | \theta - \hat{\theta} | 
\quad\text{and}\quad
\int_{K_\text{in}} \left\| f_{\theta}( \boldsymbol{x} ) - f_{\hat{\theta}} (\boldsymbol{x}) \right\|_2^2 \dif \mu(\boldsymbol{x}) \leq C_\theta \ | \theta - \hat{\theta} | 
\ ,
\]
where $C_\theta$ is a universal constant. The full proof of Theorem~\ref{thm:rate_att} can be obtained from Appendix~\ref{app:thm_rate_att}.

Next, we can derive the theoretical results relative to the expressive degradation of MLPs.
\begin{theorem}  \label{thm:rate_mlp}
Let $K_\text{h}$ be a compact set in $[-D,D]^H$ where $D>0$, $\mu$ is a probability measure defined on $K_\text{h}$, and $f_{Q_n(\theta)}( \cdot )$ is a function expressed by an MLP with a width of at most $N_w$, a depth of $L$, and the ReLU activation. For any $\epsilon > 0$, if there exists 
\[
\delta_1 = \mathcal{O} \left( C_{\textrm{mlp}}^{-1} D^{-1} n^{-L} \epsilon \right)
\quad\text{with}\quad
C_{\textrm{mlp}} = LN_w^L
\]
satisfying $\max_\theta | Q_n(\theta) - \theta | \leq \delta_1$, it holds
\[
\left\| f_{Q_n(\theta)}( \boldsymbol{x} ) - f_\theta( \boldsymbol{x} ) \right\|_{L_\mu^\infty(K_\text{h},\mathbb{R}^M)} \leq \epsilon \ .
\]
Further, if there exists $\delta_2 = \mathcal{O}( \sqrt{H} \mu(K_\text{h}) ) \cdot \delta_1$ such that $\max_\theta | Q_n(\theta) - \theta | \leq \delta_2$, it holds
\[
\left\|  f_{Q_n(\theta)}( \boldsymbol{x} ) - f_\theta( \boldsymbol{x} ) \right\|_{L_\mu^2(K_\text{h},\mathbb{R}^M)} \leq \epsilon \ .
\]
\end{theorem}
Theorem~\ref{thm:rate_mlp} shows that the approximation gap of the weight-quantized MLPs approaching real-valued ones would decrease polynomially as the number of quantization bits increases, in which $n^{-L}$ and $C_{\textrm{mlp}} = LN_w^L$ indicate the approximation effects led by weight quantization and architecture configurations, respectively. The key idea of proving Theorem~\ref{thm:rate_mlp} is similar to that of proving Theorem~\ref{thm:rate_att}. The full proof of Theorem~\ref{thm:rate_mlp} is listed in Appendix~\ref{app:thm_rate_mlp}.

\noindent\textbf{Remarks.} Since the Transformer is constructed by the ATT and MLP modules in sequence, the approximation effect of the weight-quantized Transformer is the product of those of the weight-quantized ATT and MLP modules, that is, $C_{\textrm{att}} C_{\textrm{mlp}}$ from Theorem~\ref{thm:rate_att} and Theorem~\ref{thm:rate_mlp}. This completes the proof of Theorem~\ref{thm:UA_bound_informal}. $\hfill\square$

\section{Experiments}   \label{sec:experiments}
This section conducts numerical experiments to verify the effectiveness of our theoretical results, particularly the relation between the number of quantization bits $n$, output gap $\epsilon$, weight gap $\delta$, and some factors relative to the model complexity from Theorem~\ref{thm:UA_bound_informal}.

\begin{figure}[t]
	\centering
	\begin{minipage}{0.49\linewidth}
		\centering
		\includegraphics[width=1\linewidth]{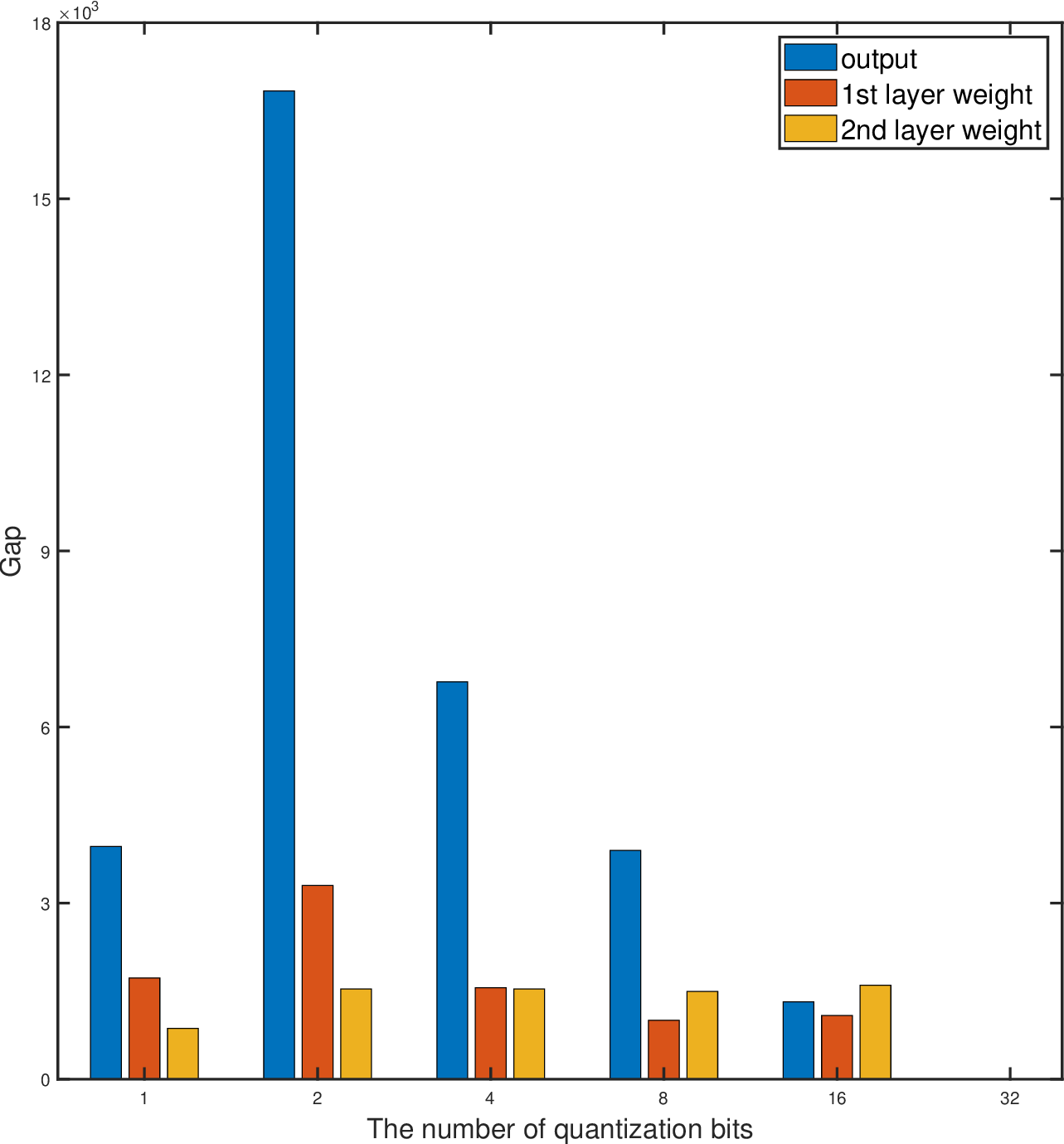}
		\centering{(a)}
	\end{minipage}
	\begin{minipage}{0.49\linewidth}
		\centering
		\includegraphics[width=1\linewidth]{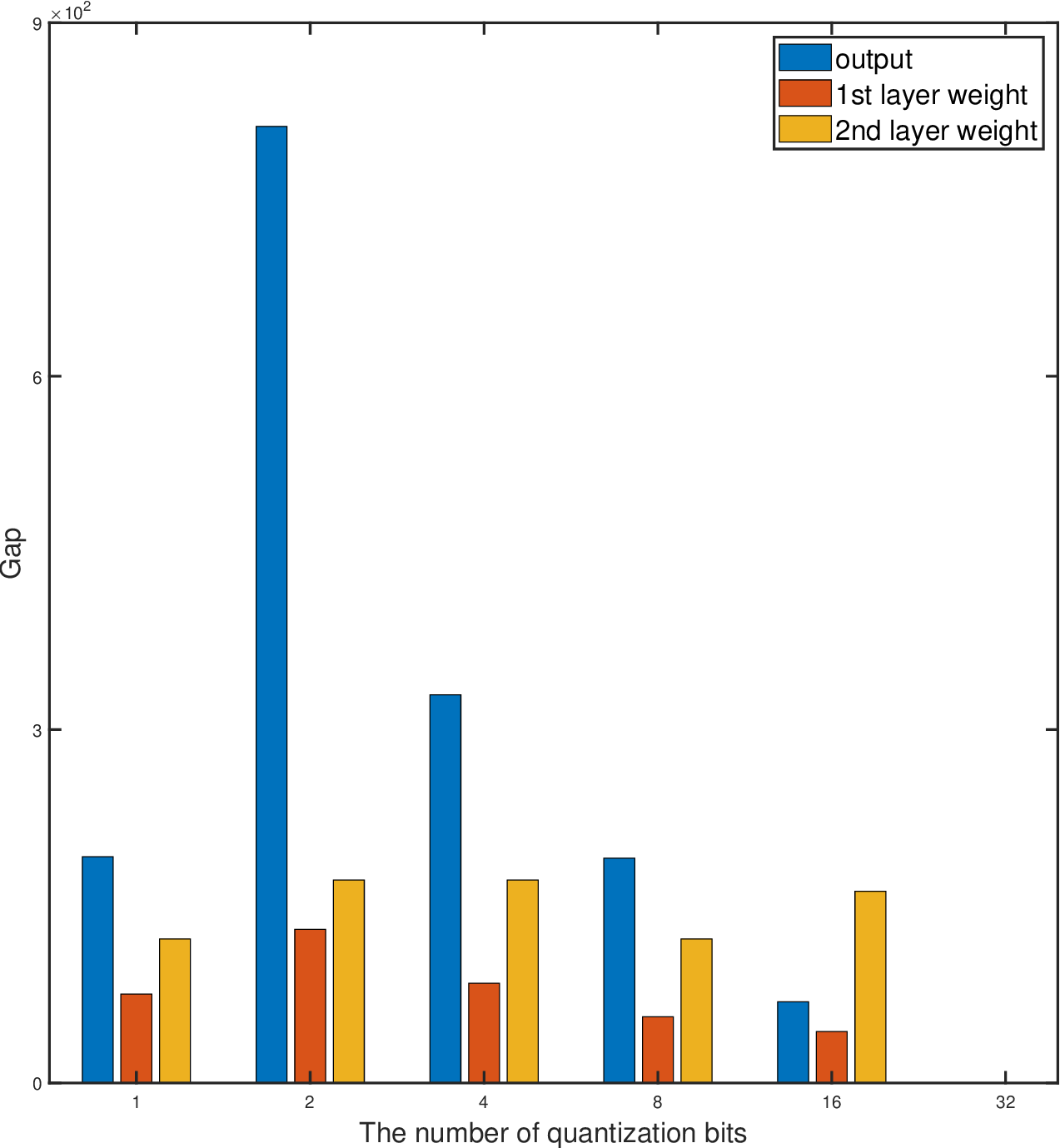}
		\centering{(b)}
	\end{minipage}
	\caption{The approximation gap bars of weight-quantized MLP within (a) 2-norm and (b) $\infty$-norm with respect to the number of quantization bits $n$.} 
	\label{fig:bar}
\end{figure}

\subsection{Simulation Regression of Weight-Quantized MLPs} \label{subsec:regression}
The first experiment conducts the regression task on simulation data. We generate 1000 pairs of simulation points from $([-1,1]^{10}, \mathbb{R})$ and employ a one-hidden-layer MLP with 100 hidden neurons. In these settings, the factors related to model complexity are fixed accordingly, i.e., $D=1$, $L=1$, $N=10$, $N_w = 100$, and $M=1$. Besides, we employ the typical floating as a baseline, that is, $n=32$, and take the number of quantization bits from $\{2^0, 2^1, 2^2, 2^3, 2^4, 2^5\}$. The above configurations meet the conditions of Theorem~\ref{thm:rate_mlp}. By exploiting the relation among $n$, $\delta$, and $\epsilon$, we can verify the explicit bound, especially the order of magnitude function $\mathcal{O}(\cdot)$ in Theorem~\ref{thm:rate_mlp}.

Figure~\ref{fig:bar} plots approximation gap bars of the output, 1st-layer weight, and 2nd-layer weight with respect to the number of quantization bits within norms of $\| \cdot \|_2$ and $\| \cdot \|_\infty$, which are marked as 2-norm and $\infty$-norm in the figures, respectively. For visibility, we scale the 2-norms of weight gaps by a factor of 100 and the $\infty$-norm of the 2nd layer weight gap by 50. It is observed that neither the relation between $\epsilon$ and $n$ nor between $\delta$ and $n$ is explicit.

\begin{figure}[t]
	\centering
	\begin{minipage}{0.49\linewidth}
		\centering
		\includegraphics[width=1\linewidth]{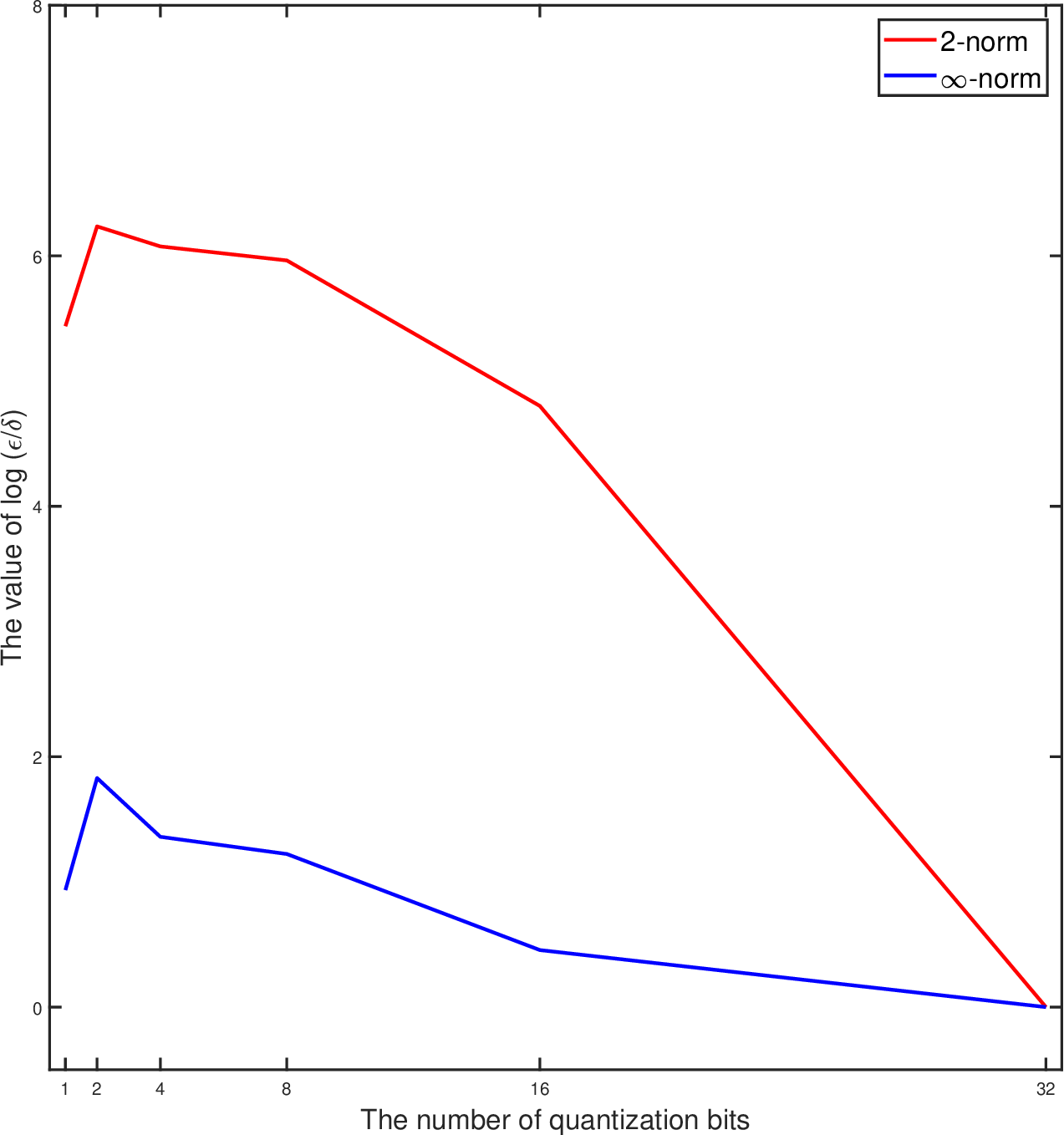}
		\centering{(a)}
	\end{minipage}
	\begin{minipage}{0.49\linewidth}
		\centering
		\includegraphics[width=1\linewidth]{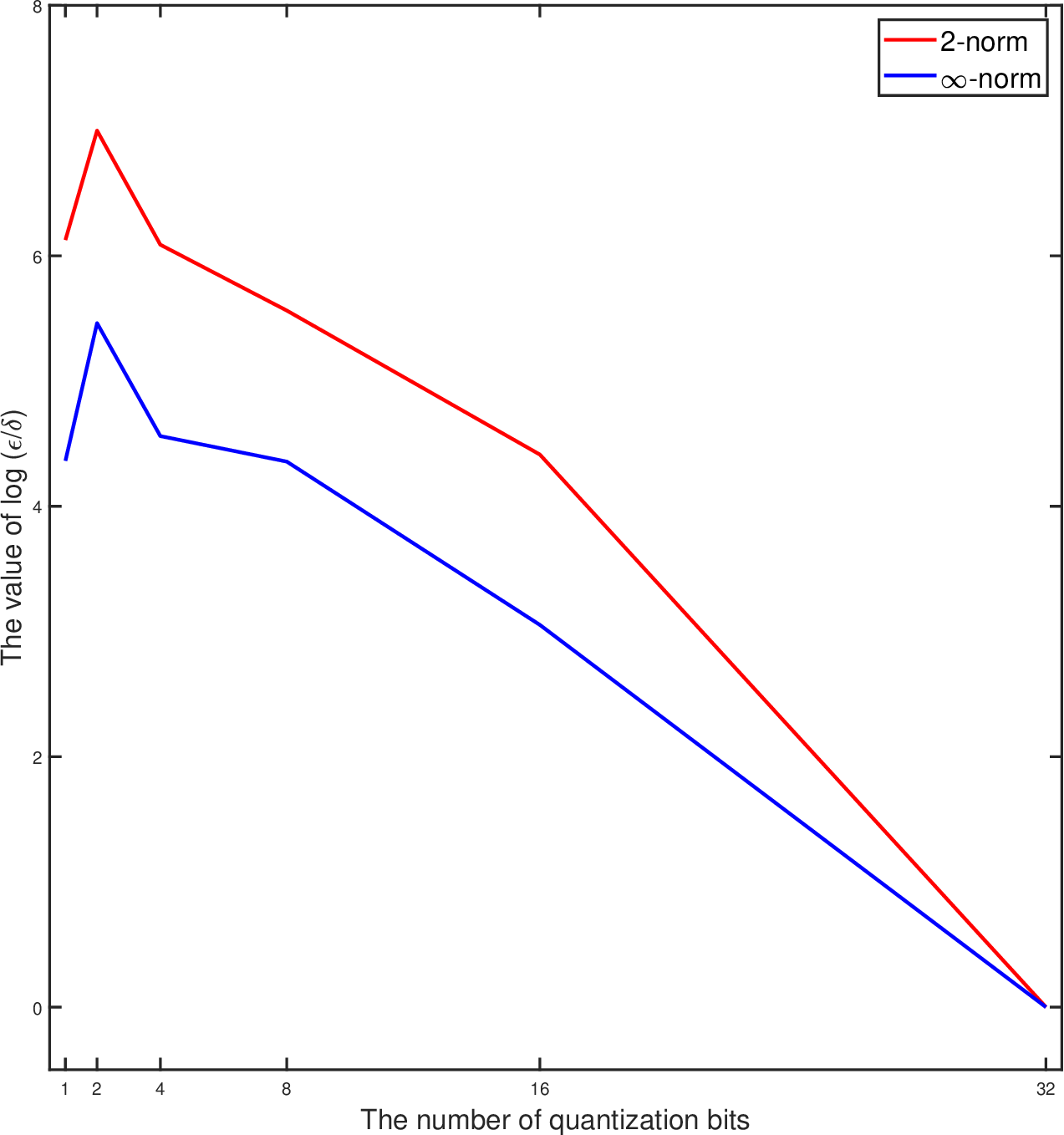}
		\centering{(b)}
	\end{minipage}
	\caption{The relation curves between the ratio $\ln (\epsilon / \delta)$ and the number of quantization bits $n$ within (a) 2-norm and (b) $\infty$-norm.} 
	\label{fig:curves}
\end{figure}

Furthermore, we define an indicator $\ln (\epsilon / \delta)$; ideally, there exists a negative correlation between the indicator $\ln (\epsilon / \delta)$ and the number of quantization bits $n$ according to Theorem~\ref{thm:rate_mlp}. Figure~\ref{fig:curves} displays the empirical relation between $n$ and $\ln (\epsilon / \delta)$ within the 2-norm and $\infty$-norm, respectively. These two figures signify that $\ln (\epsilon / \delta)$ is inversely proportional to $n$, i.e., $\ln (\epsilon / \delta) \leq a n + b$, where $a<0$ and $b>0$. These observations hold regardless of whether the Euclidean or infinite norm is used, confirming the effectiveness of Theorem~\ref{thm:rate_mlp}.

\subsection{ImageNet Classification of Weight-Quantized Deep Learning Models}  \label{subsec:classification}
\begin{figure}[t]
	\centering
	\begin{minipage}{0.49\linewidth}
		\centering
		\includegraphics[width=1\linewidth]{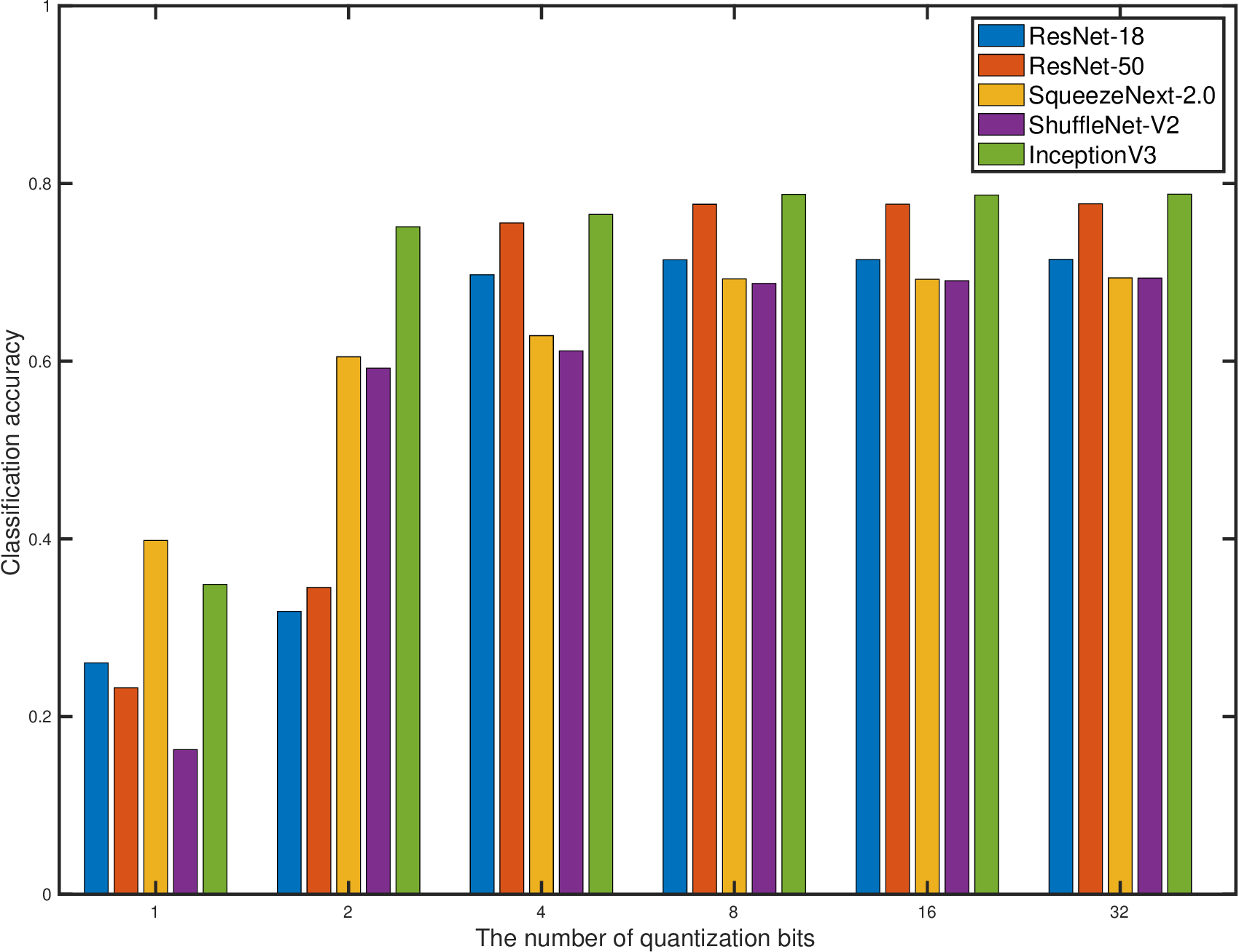}
		\centering{(a)}
	\end{minipage}
	\begin{minipage}{0.49\linewidth}
		\centering
		\includegraphics[width=1\linewidth]{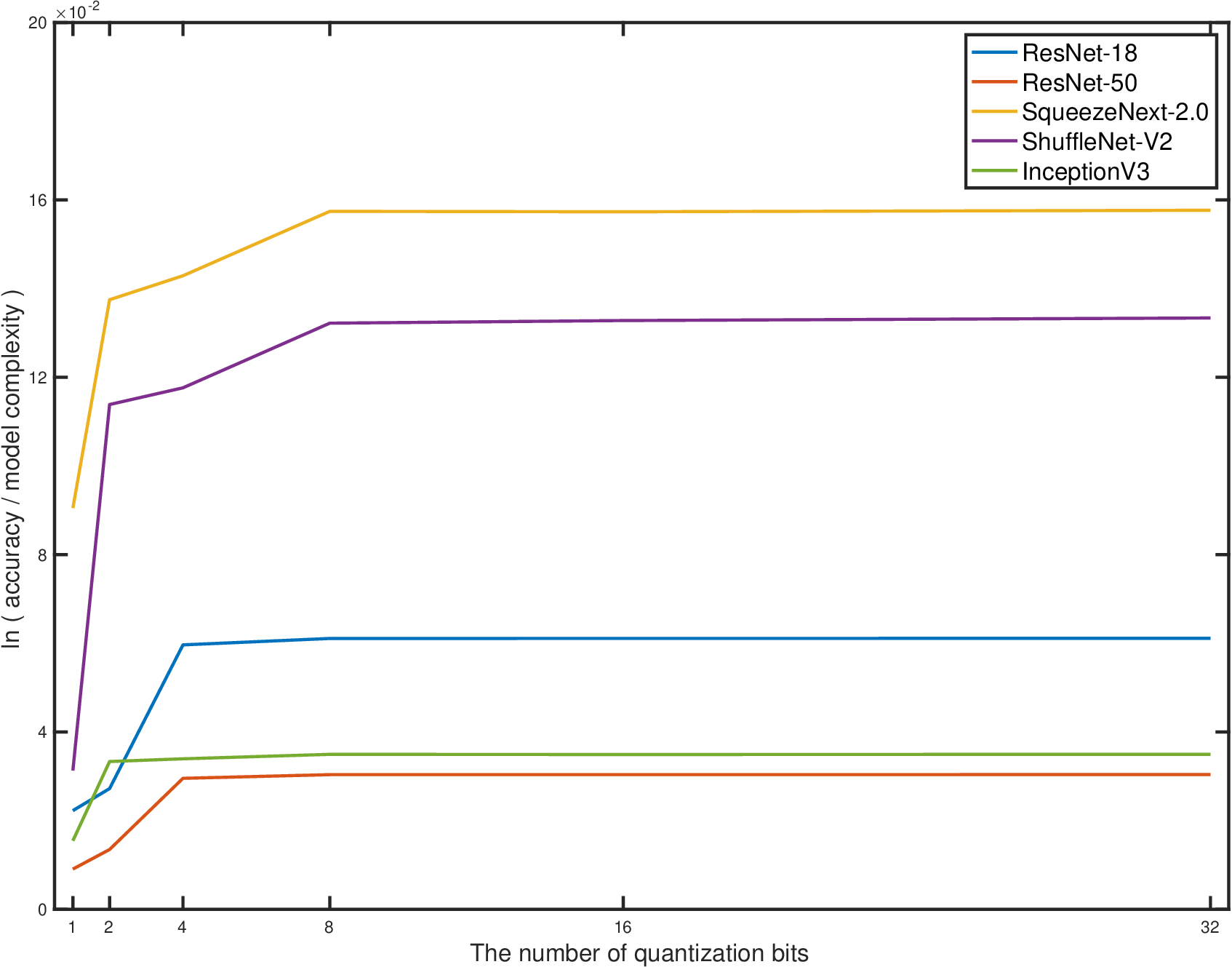}
		\centering{(b)}
	\end{minipage}
	\caption{(a) The accuracy bars of conducted models with respect to the number of bits $n$ and (b) the relation curves between the number of quantization bits $n$ and $\ln ( \text{accuracy} / \text{model complexity} ) $. }
	\label{fig:acc-paras}
\end{figure}

The second experiment conducts the classification tasks on the ImageNet dataset~\citep{deng2009imagenet}. The investigated models contain ResNet-18, ResNet-50~\citep{he2016:resnet}, SqueezeNext~\citep{ma2018:shufflenet}, ShuffleNet-V2~\citep{szegedy2016:inception}, and Inception-V3~\citep{iandola2018:squeezenext}. We conduct experiments on NVIDIA RTX 6000 Ada $\times$ 8 and implement each model with different quantization bits. Figure~\ref{fig:acc-paras} plots the accuracy of the investigated models as the number of quantization bits varies from $\{2^0, 2^1, 2^2, 2^3, 2^4, 2^5\}$. Figure~\ref{fig:acc-paras}(a) shows the accuracy of the conducted models with respect to the number of bits. It is obvious that all models with 1-bit quantization maintain poor performance.

Since these models maintain different architectures, it is not easy to compute factors relative to model complexity; instead, we count the number of parameters as an indicator of model complexity. In addition, it is also challenging to compute the weight gap $\delta$ layer by layer; in other words, $\delta$ is unknown. Thus, the results of Theorem~\ref{thm:rate_mlp} become the relation among $n$, $\epsilon$, and model complexity. Here, we replace $\epsilon$ by classification accuracy and present an indicator $\ln ( \text{accuracy} / \text{model complexity} )$, e.g., 8-bit ResNet-18 takes $\ln (71.43 / 11.69)$. From Theorem~\ref{thm:rate_mlp}, there ideally exists a positive correlation between this indicator and the number of quantization bits. Figure~\ref{fig:acc-paras}(b) plots the empirical relation between $n$ and $\ln ( \text{accuracy} / \text{model complexity} ) $. It is evident that the indicator $ \ln ( \text{accuracy} / \text{model complexity} ) $ is proportional to the number of quantization bits $n$, i.e., $\ln ( \text{accuracy} / \text{model complexity} ) \leq a n + b$, where $a>0$ and $b>0$, which demonstrates the effectiveness of our theoretical results.

\subsection{WikiText2 Perplexity of Weight-Quantized LLMs}  \label{subsec:wiki}
The third experiment is carried out on the WikiText2 dataset~\citep{merity2016:wiki} with the perplexity metric. The conducted models comprise MobileLLM-125M/350M/600M/1B/1.5B~\citep{liu2024:mobilellm} and LLaMA-1B/3B/8B~\citep{Llama3}. All Weight-Quantized models are trained with LLM-QAT~\citep{liu2023:llmqat}, where all weights except for the embedding and output layers are quantized. For weight-quantized model training, we initialize the models with pre-trained weights and employ the AdamW optimizer~\citep{loshchilov2017:adamw} with zero weight decay and a batch size of 8. Further, the optimization process spanned $1.2 \times 10^{5}$ iterations with an initial learning rate of $2 \times 10^{-5}$ for the cases of 1-bit and 2-bit quantization, while the process involved $4 \times 10^4$ iterations with an initial learning rate of $1 \times 10^{-5}$ for the cases of 4-bit, 8-bit, and 16-bit quantization. The learning rate decayed to zero following a cosine learning rate decay. For the training with floating numbers, it is adopted with a batch size of 16, a weight decay of 0.1, an initial learning rate of $2.5 \times 10^{-3}$, and a linear learning rate decay to zero. We conduct experiments on NVIDIA A100 80GB $\times$ 8 $\times$ 2 and implement each model with the number of quantization bits varying from $\{2^0,2^1,2^2,2^3,2^4\}$. 

\begin{figure}[t]
	\centering
	\begin{minipage}{0.325\linewidth}
		\centering
		\includegraphics[width=1\linewidth]{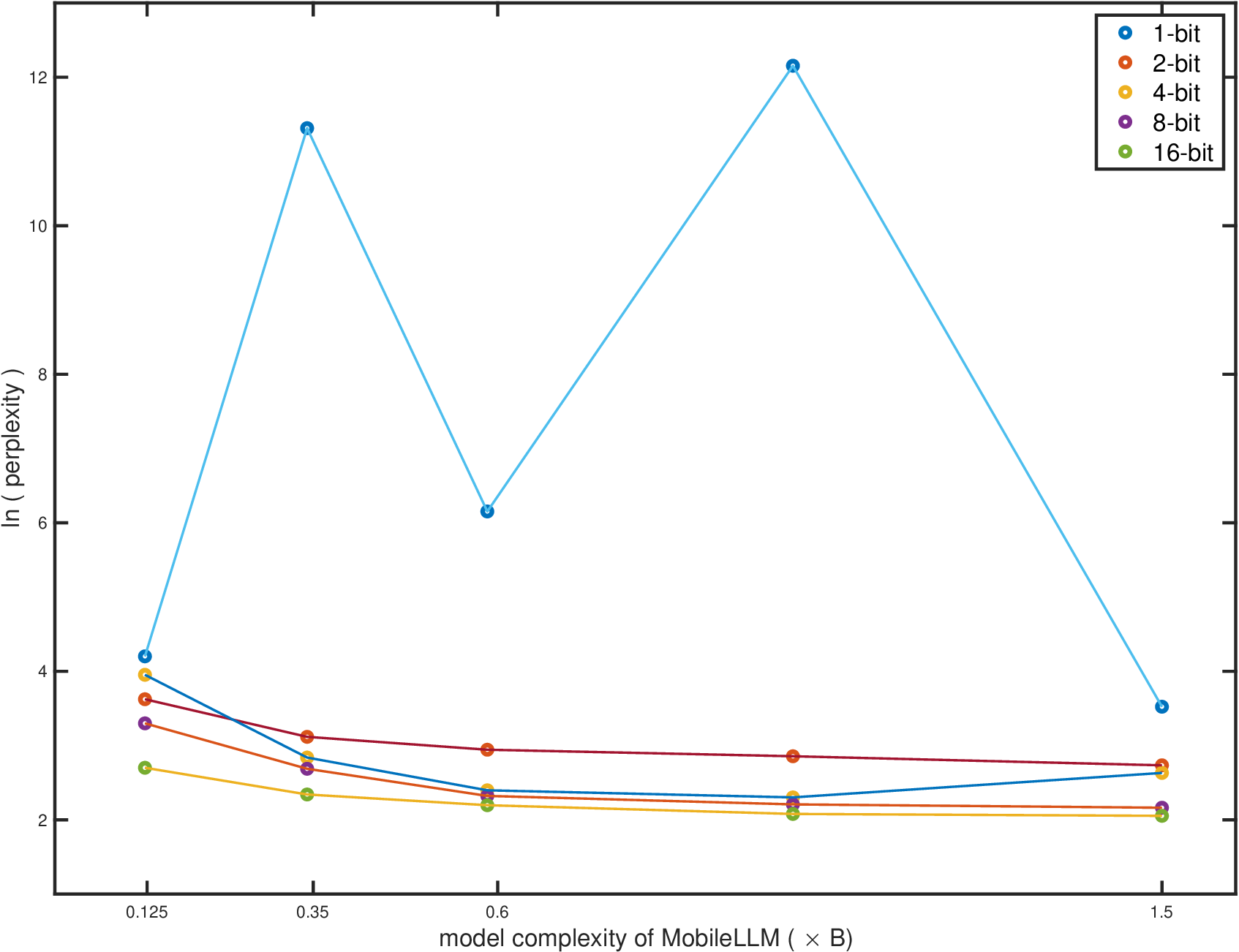}
		\centering{(a)}
	\end{minipage}
	\begin{minipage}{0.325\linewidth}
		\centering
		\includegraphics[width=1\linewidth]{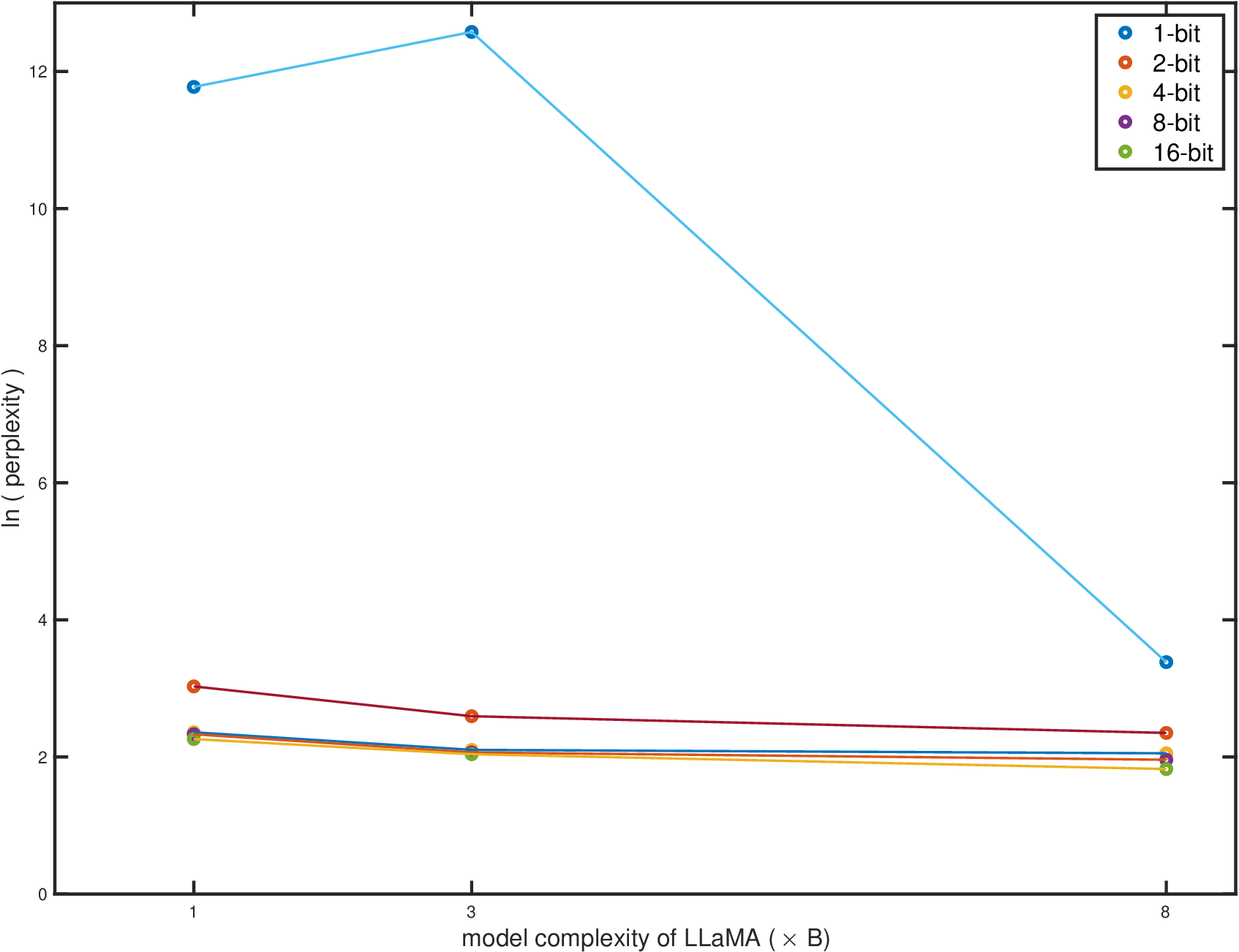}
		\centering{(b)}
	\end{minipage}
	\begin{minipage}{0.325\linewidth}
		\centering
		\includegraphics[width=1\linewidth]{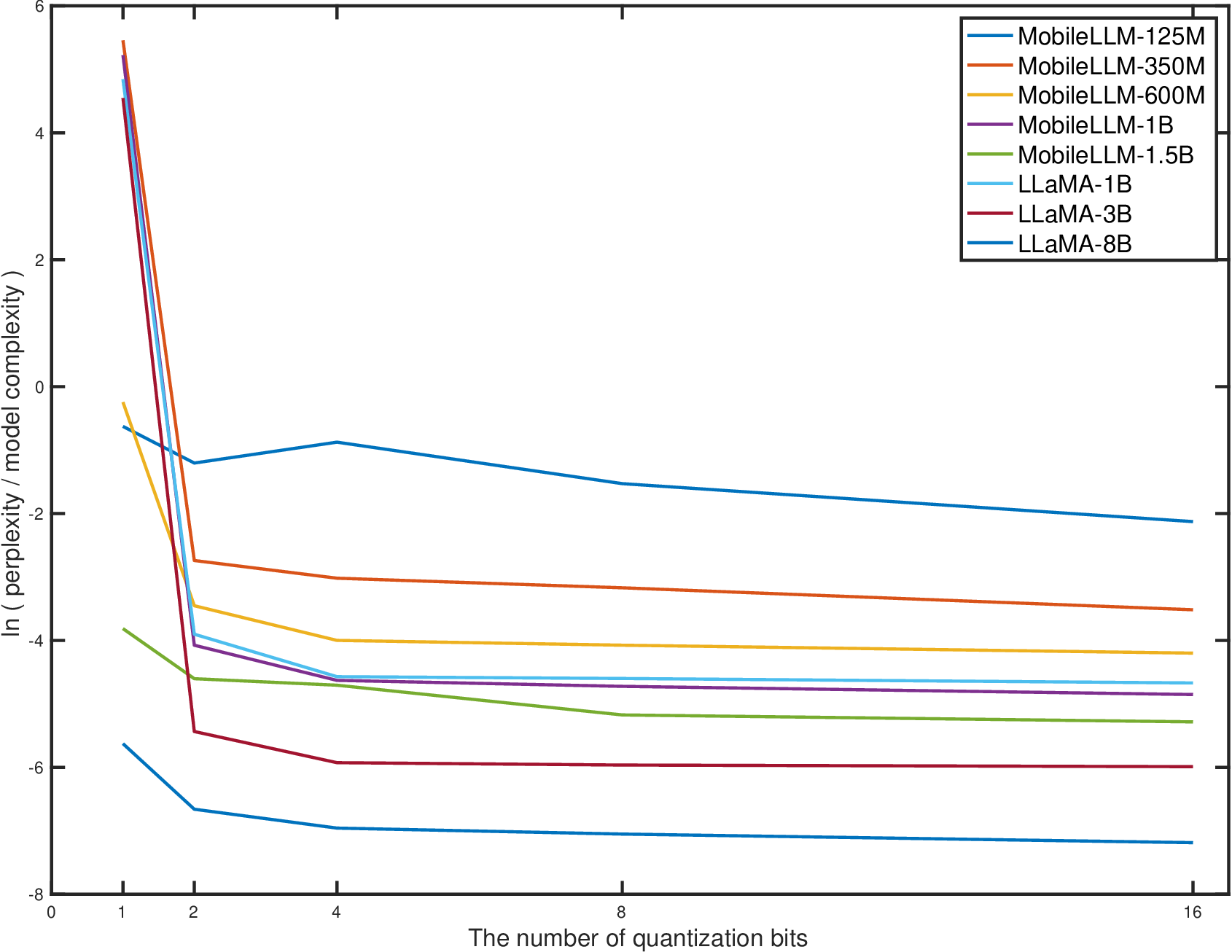}
		\centering{(c)}
	\end{minipage}
	\caption{The relation curves (a-b) between the model complexity and $\ln ( \text{perplexity} )$ and (c) between the number of quantization bits $n$ and $\ln ( \text{perplexity} / \text{model complexity} ) $ of weight-quantized MobileLLMs and LLaMAs.} 
	\label{fig:wiki}
\end{figure}

Here, we employ the indicator  $\ln ( \text{perplexity} / \text{model complexity} ) $. Figure~\ref{fig:wiki}(c) plots the relation curves between the number of quantization bits $n$ and $\ln ( \text{perplexity} / \text{model complexity} ) $ of MobileLLMs and LLaMAs. There are two observations: (1) In the case of $[2,4,8,16]$ quantization bits, the indicator $ \ln ( \text{perplexity} / \text{model complexity} ) $ is proportional to the number of quantization bits $n$, i.e., $\ln ( \text{perplexity} / \text{model complexity} ) \leq a n + b$, where $a \lesssim 0$ and $b>0$, confirming the effectiveness of Theorem~\ref{thm:UA_bound_informal}; (2) Those of the 1-bit case have poor performance and show the sudden change from the aforementioned near-linear bound, verifying the effectiveness of Theorem~\ref{thm:1-bit_mlp}. Figure~\ref{fig:wiki}(a-b) displays the relation curves between the model complexity and $\ln ( \text{perplexity} )$ in detail.

\subsection{Zero-shot Reasoning of Weight-Quantized LLMs}  \label{subsec:reasoning}
The fourth experiment inherits the conducted LLMs and the corresponding configurations in Subsection~\ref{subsec:wiki}, with performance evaluation across eight zero-shot common sense reasoning tasks, including ARC-easy, ARC-challenge~\citep{clark2018:ARC}, BoolQ~\citep{clark2019:boolq}, PIQA~\citep{bisk2020:piqa}, Social-IQA~\citep{sap2019:socialiqa}, HellaSwag~\citep{zellers2019:hellaswag}, OBQA~\citep{mihaylov2018:OBQA}, and WinoGrande~\citep{sakaguchi2021:winogrande}. Table~\ref{tab:zero_shot} lists the accuracy of MobileLLMs and LLaMAs equipped with 2-, 3-, and 4-bit weights on zero-shot reasoning tasks, the observations of which coincide with those of the aforementioned experiments.
\begin{table}[t]
	\caption{The accuracy of weight-quantized MobileLLMs and LLaMAs with formatting weights into 2, 3, and 4 bits on zero-shot reasoning tasks.}
	\label{tab:zero_shot}
	\resizebox{\textwidth}{!}{
	\begin{tabular}{c|c|c|c|c|c|c|c|c|c|c}
		\toprule
		\textbf{Models}                 & \textbf{Bits} & \textbf{ARC-easy} & \textbf{ARC-challenge} & \textbf{BoolQ} & \textbf{PIQA} & \textbf{Social-IQA} & \textbf{HellaSwag} & \textbf{OBQA} & \textbf{WinoGrande} & \textbf{AVERAGE} \\ \midrule
		\multirow{3}{*}{MobileLLM-125M} & 2             & 34.9              & 23.3                   & 61.8           & 53.8          & 39.3                & 29.1               & 27.4          & 51.3                & 40.1             \\ \cline{2-11} 
		& 3             & 44.7              & 28.7                   & 53.7           & 60.6          & 41.1                & 34.6               & 34.9          & 50.2                & 43.5             \\ \cline{2-11} 
		& 4             & 54.2              & 33.4                   & 52.2           & 64.7          & 42.4                & 39                 & 42.7          & 51.7                & 47.5             \\ \midrule
		\multirow{3}{*}{MobileLLM-350M} & 2             & 40.6              & 25.9                   & 62             & 55.6          & 40                  & 31.8               & 31.1          & 52.6                & 42.5             \\ \cline{2-11} 
		& 3             & 54.6              & 35.4                   & 60.5           & 65.9          & 42.2                & 42.6               & 41.9          & 53.4                & 49.5             \\ \cline{2-11} 
		& 4             & 63.4              & 42                     & 59.8           & 70.1          & 43.6                & 51.5               & 47            & 57.5                & 54.4             \\ \midrule
		\multirow{3}{*}{MobileLLM-600M} & 2             & 42.7              & 25.6                   & 62.1           & 56            & 38.8                & 33.7               & 29.6          & 51.5                & 42.5             \\ \cline{2-11} 
		& 3             & 61.8              & 38                     & 62.1           & 68.5          & 43.6                & 48.9               & 44.2          & 54.6                & 52.7             \\ \cline{2-11} 
		& 4             & 67.2              & 47.4                   & 65.2           & 71.8          & 43.8                & 57.8               & 50.6          & 59.8                & 57.9             \\ \midrule
		\multirow{3}{*}{MobileLLM-1B}   & 2             & 42.6              & 26.7                   & 49.7           & 57.7          & 40.4                & 34.9               & 31.4          & 49.2                & 41.6             \\ \cline{2-11} 
		& 3             & 65.3              & 42.6                   & 61.2           & 70.4          & 44                  & 54.3               & 48.8          & 56.8                & 55.5             \\ \cline{2-11} 
		& 4             & 72.1              & 49.5                   & 66.1           & 73.9          & 46.2                & 63                 & 55.4          & 63.7                & 61.2             \\ \midrule
		\multirow{3}{*}{MobileLLM-1.5B} & 2             & 45.3              & 26.5                   & 61.6           & 58.6          & 40.1                & 37.5               & 33.1          & 50.6                & 44.2             \\ \cline{2-11} 
		& 3             & 68.6              & 44.4                   & 62.4           & 71.8          & 45.4                & 57.8               & 49.2          & 57.2                & 57.1             \\ \cline{2-11} 
		& 4             & 72.3              & 49.5                   & 70.1           & 73.5          & 47.1                & 64.5               & 53.2          & 63.4                & 61.7             \\ \midrule
		\multirow{3}{*}{LLaMA-1B}       & 2             & 49.2              & 33.3                   & 62             & 63.9          & 41.1                & 41.5               & 37.5          & 54.4                & 47.9             \\ \cline{2-11} 
		& 3             & 52.7              & 32.4                   & 60.5           & 66.6          & 44                  & 43.2               & 40.2          & 53.8                & 49.2             \\ \cline{2-11} 
		& 4             & 61.3              & 38.1                   & 62.3           & 73            & 44.2                & 59                 & 41.8          & 58.7                & 54.8             \\ \midrule
		\multirow{3}{*}{LLaMA-3B}       & 2             & 49.3              & 33.3                   & 63.5           & 65.2          & 41.7                & 48.9               & 34.2          & 52.2                & 48.5             \\ \cline{2-11} 
		& 3             & 64.4              & 40.1                   & 62             & 71.7          & 45                  & 58.2               & 44.7          & 59.9                & 55.8             \\ \cline{2-11} 
		& 4             & 71.8              & 48.1                   & 74.6           & 76.6          & 48.1                & 71.4               & 52.3          & 67.4                & 63.8             \\ \midrule
		\multirow{3}{*}{LLaMA-8B}       & 2             & 54.8              & 35.9                   & 64.8           & 68            & 41.8                & 58                 & 35.7          & 54.7                & 51.7             \\ \cline{2-11} 
		& 3             & 68.8              & 48.8                   & 71.1           & 75.9          & 46.8                & 67.8               & 48.2          & 65.1                & 61.6             \\ \cline{2-11} 
		& 4             & 77.4              & 54                     & 82.9           & 79.1          & 49.2                & 77.6               & 54.3          & 72                  & 68.3             \\   \bottomrule
	\end{tabular} }
\end{table}

\section{Conclusions and Prospects}  \label{sec:conclusions}
In this paper, we theoretically investigated the expressive capacity of LLMs equipped with weight quantization. We concluded three main results: (1) We established the universal approximation property of weight-quantized LLMs with linear width and exponential depth when the number of quantization bits is greater than 1; (2) we argued that the ternary format, or equally 1.58-bit, is the limiting precision for weight quantization, as there is a counterexample function living on $[-1,1]^N$ that cannot be uniformly approximated by 1-bit weight-quantized models; (3) we confirmed that weight quantization leads to expressive degradation, in which the expressive power of weight-quantized LLMs degrades polynomially as the number of quantization bits decreases. Comprehensive experiments demonstrated the effectiveness of the three theoretical results mentioned above.

Our theoretical results provide a solid foundation for advancing weight quantization in the context of scaling laws and shed insights for future research in model compression and inference acceleration. In the future, it is significant to develop in-depth theoretical understandings of weight quantization from aspects of computational efficiency and generalization ability within over-parameterized architectures. It is also attractive to explore the theoretical analysis of activation quantization. In addition, it is worth discussing the theoretical characterizations of foundation models with quantization and developing the algorithms that combine quantization with scaling law, although universal approximation builds upon considerably larger architectures.

\section*{Acknowledgements}
This research was supported by the National Science Foundation of China (62406138).

\appendix
\section{Full Proof of Lemma~\ref{lemma:hierarchical_mlp}} \label{app:lemma_hierarchical_mlp}

As stated with the ReLU activation, it is observed that
\[
x = \left\{\begin{aligned}
	1 \cdot \sigma \left( 1 \cdot x  + 0 \right) \ , & \quad\text{for}\quad x \geq 0 \ , \\
	\omega \cdot \sigma \left( \omega \cdot x + 0 \right) \ , & \quad\text{for}\quad x \leq 0 \ .
\end{aligned}\right.
\]
Thus, the concerned identity function $\mathcal{I}: \mathbb{R} \to \mathbb{R}$ can be approximated well by a $n$-bit weight-quantized MLP with two layers and a width of 1 according to $\mathcal{I}(x) = q \cdot \sigma ( q \cdot x  + 0 )$, where $q = 1$ for $x \geq 0$ and $q = \omega$ for $x <0$.

For the square one, we have the following decomposition
\[
x^2 = \sum_{i=1}^x (2i -1)  \quad\text{for}\quad  x \in \mathbb{Z} \ .
\]
Thus, the concerned identity function $\mathcal{S}: \mathbb{R} \to \mathbb{R}$ can be approximated well by a $n$-bit weight-quantized MLP with two layers and a width of $\mathcal{O}(x)$ according to
\[
\mathcal{S}(x) = \sum_{i=1}^x \mathcal{I} \left( \Delta(i) + \omega \cdot 1 \right) 
\quad\text{with}\quad
\Delta(x) = \mathcal{I} (x) + \mathcal{I} (x)
\quad\text{for}\quad
n\geq 2 \ .
\]
For the case of $n=1$, Algorithm~\ref{alg:bit_wise_fcn} displays the computing procedure. 

For the multiplication, it is observed that
\[
x_1 x_2 = \frac{1}{2^2} \left(  (x_1+x_2)^2 - (x_1-x_2)^2  \right) \ .
\]
Thus, the concerned identity function $\mathcal{M}: \mathbb{R} \to \mathbb{R}$ can be approximated well by a $n$-bit weight-quantized MLP with two layers and a width of $\mathcal{O}(x)$ according to 
\[
\mathcal{M}(x) = \left[ \mathcal{S}( x_1 + x_2 ) -  \mathcal{S}( x_1 - x_2 ) \right] \gg 2
\quad\text{for}\quad
n\geq 2 \ .
\]
For the case of $n=1$, one can compute $\mathcal{M}(x)$ from Algorithm~\ref{alg:bit_wise_fcn}. This completes the proof of $(3)$.

For the reciprocal, one has
\[
\frac{1}{x} = \lim\limits_{m \to \infty} (2-x) \prod_{i=1}^{m} \left[  1 + (1-x)^{2^i} \right] 
\quad\text{and thus}\quad
\mathcal{R}(x) = \lim\limits_{m \to \infty} \mathcal{I}(2-x) 
\prod_{i=1}^{m} \mathcal{I} \left[  1 + \mathcal{S}^i(1-x) \right] \ .
\]
The concerned reciprocal function $\mathcal{R}: \mathbb{R} \to \mathbb{R}$ can be uniformly approximated by a $n$-bit weight-quantized MLP with two layers according to Algorithm~\ref{alg:bit_wise_fcn}. This completes the proof. $\hfill\square$

\section{Finishing the Proof of Theorem~\ref{thm:UA_mlp}} \label{app:thm_UA_mlp}
Let $\boldsymbol{f} \in \mathcal{C}(K_\text{h}, \mathbb{R}^M)$, where $\boldsymbol{f} = (f_1, \dots,f_M)$. Decomposing a $n$-bit weight-quantized MLP into $M$ sub-networks that correspond to $M$ target sub-functions, the problem of approximating $\boldsymbol{f}$ can be degenerated to another problem of approximating $f_j$ for any $j\in[M]$. It suffices to enable each approximation to the target sub-function within error $\epsilon>0$ on $K_\text{h}$.

It is evident that any continuous function can be approximated well by a polynomial function, which comprises three types of hierarchical operations, i.e., the identity, upgrade, and degrade operations. Let $\boldsymbol{x} \in \mathbb{R}^H$ denote the inputs, where $\boldsymbol{x} = (x_1, \dots, x_H)$.  Each hidden layer has three groups of neurons. The first group of neurons simply records the input $(x_1, \dots, x_H)$ by applying an identity activation function, in requirement of $H$ hidden neurons from Lemma~\ref{lemma:hierarchical_mlp}. The second group of neurons performs its computation based on the inputs preserved in other hidden neurons, in requirement of one hidden neuron. The third group of neurons is used to compute these hierarchical operations. According to Lemma~\ref{lemma:hierarchical_mlp}, the number of hidden neurons is at most $\max\{x_1, \dots,x_H\}$, that is, $\mathcal{O}(\|\boldsymbol{x}\|_{\infty})$. Hence, each hidden layer has $H+1+\mathcal{O}(\|\boldsymbol{x}\|_{\infty})$ neurons arranged into a group of $H$ neurons, a group of a single neuron, and a group of $\mathcal{O}(\|\boldsymbol{x}\|_{\infty})$ neurons.  Each approximation to the target sub-function is within error $\epsilon>0$ on $K_\text{h}$.

Finally, the neurons in the output layer of the network are connected to the final hidden layer and equipped with the identity activation function as usual. Uniform continuity preserves uniform convergence, continuous functions preserve compactness, and a composition of two uniformly convergent sequences of functions with uniformly continuous limits is again uniformly convergent. Thus, we can combine $M$ sub-networks, which approximate the target sub-functions within error $\epsilon>0$ on $K$, by taking a sufficiently larger number of hidden layers. Thus, the width of each hidden layer is at most $H+M+\mathcal{O}(\|\boldsymbol{x}\|_{\infty})$. This proof is finished. $\hfill\square$

\section{Full Proof of Lemma~\ref{lemma:hierarchical_att}}  \label{app:lemma_hierarchical_att}

For convenience, we here consider the 1-dimensional output, that is, $H=1$. Thus, Eq.~\eqref{eq:conversion} becomes
\[
\tilde{z} = \sum_{i=1}^{N_h} \text{ReLU}\left[ \left(\boldsymbol{w}_{\!\text{k}}^{(i)} \boldsymbol{x} + b_\text{k}^{(i)} \right)^\top \left( \boldsymbol{w}_{\!\text{q}}^{(i)} \boldsymbol{x} + b_\text{q}^{(i)} \right) \right]  \left( \boldsymbol{w}_\text{o}^{(i)} \odot \boldsymbol{w}_\text{v}^{(i)} \ \boldsymbol{x} + \boldsymbol{w}_\text{o}^{(i)} \boldsymbol{b}_\text{v}^{(i)} \right) + b_\text{o}^{(i)} \ ,
\]
where $\boldsymbol{w}_{\!\text{k}}^{(i)}, \boldsymbol{w}_{\!\text{q}}^{(i)}, \boldsymbol{w}_{\!\text{o}}^{(i)}, \boldsymbol{w}_{\!\text{v}}^{(i)}$ are raw vectors and $\odot$ denotes the cross product. As stated with the ReLU activation, it suffices to prove that
\[
\begin{aligned}
\mathcal{I}(x) &= \text{ReLU}\left[ ( 0 \cdot x + 1 ) ( 0 \cdot x + 1 )  \right] (1 \cdot 1 \cdot x + 1 \cdot 0) + 0 \\
\mathcal{S}(x) &= \text{ReLU}\left[ ( 1 \cdot x + 0 ) ( 1 \cdot x + 0 )  \right] (1 \cdot 0 \cdot x + 1 \cdot 1) + 0 \\
\mathcal{M}(x_1,x_2) &= \text{ReLU}\left[ ( (1,0) \cdot (x_1,x_2)^\top + 0 ) ( (0,q) \cdot (x_1,x_2)^\top + 0 )  \right] (1 \cdot 0 \cdot x + 1 \cdot q) + 0  \\
\mathcal{R}(x) &= \lim\limits_{m \to \infty} (2-x) \prod_{i=1}^{m} \left[  1 + (1-x)^{2^i} \right]
= \lim\limits_{m \to \infty} \mathcal{I}(2-x) 
\prod_{i=1}^{m} \mathcal{I} \left[  1 + \mathcal{S}^i( \mathcal{I}(1-x)) \right] \ ,
\end{aligned}
\]
where $q = 1$ for $x_1x_2 \geq 0$ and $q = \omega$ for $x_1x_2 <0$. Therefore, the identity, square, and multiplication functions can be approximated well by the one-layer weight-quantized ATT with $N_h=1, N_v=1, N_k\leq 2$, while the reciprocal function can be uniformly approximated by a weight-quantized ATT with $N_h=1, N_v=1, N_k=1$ and a considerably deep architecture. This completes the proof. $\hfill\square$

\section{Full Proof of Theorem~\ref{thm:1-bit_mlp}} \label{app:1-bit_mlp}
We begin this proof with the 1-bit MLPs with 0 or 1 weights. Let $x_i$ denote the $i$-th element of $\boldsymbol{x} \in [-1,1]^H$ for $i \in [N]$. Here, we construct a counterexample function $f: [-1,1]^H \to \mathbb{R}$
\[
f(\boldsymbol{x}) = \left\{ \begin{aligned}
	\exp\left(  \frac{cx_1^2}{ cx_1^2 +1 }  \right) \ , &\quad |x_1| \leq  0.5 \ , \\
	0 \qquad\quad , &\quad |x_1| > 0.5 \ ,
\end{aligned}
\right.
\]
where $c<0$. It is observed that $f$ is a non-negative function and maintains its maximum at the original point $\boldsymbol{x} = \boldsymbol{0}$. Notice that the counterexample function $f$ cannot be exactly approximated by $f_{\textrm{1-bit}}$ but can be approximated well by real-valued MLPs.

Provided the counterexample function, we intuitively have
\begin{equation}  \label{eq:1-bit_mlp_01}
	\sup_{\boldsymbol{x} \in [-1,1]^H} \| f(\boldsymbol{x}) - f_{\textrm{1-bit}}(\boldsymbol{x}) \|
	\geq \max_{\boldsymbol{x} \in \{-1,0,1\}^H} | f(\boldsymbol{x}) - f_{\textrm{1-bit}}(\boldsymbol{x}) | \ .    
\end{equation}
This inequality in Eq.~\eqref{eq:1-bit_mlp_01} is satisfied regardless of the explicit definitions of ATTs and MLPs, as it is a direct consequence of set inclusion and supremum properties. Notice that we have the following observations on the right side of Eq.~\eqref{eq:1-bit_mlp_01}
\begin{equation}  \label{eq:1-bit_mlp_02}
| f(\boldsymbol{x}) - f_{\textrm{1-bit}}(\boldsymbol{x}) | = \left\{~\begin{aligned}
	| 1 - f_{\textrm{1-bit}}(\boldsymbol{x}) | ~, & \quad x_1 = 0 \ , \\
	| 0 - f_{\textrm{1-bit}}(\boldsymbol{x}) | ~, & \quad x_1 \in \{-1,1\} \ ,
\end{aligned} \right.
\end{equation}
where $f_{\textrm{1-bit}}(\boldsymbol{x}) \in \{-1,0,1\}$ for $\boldsymbol{x} \in \{-1,0,1\}^H$ and $f_{\textrm{1-bit}}(\boldsymbol{0})=0$. Hence, we can derive $| f(\boldsymbol{x}) - f_{\textrm{1-bit}}(\boldsymbol{x}) | \in [0,2]$. Notice that the results of Eq.~\eqref{eq:1-bit_mlp_02} also hold in the case of ATTs and MLPs that format weights into $\{-1,1\}$. Therefore, we can rewrite Eq.~\eqref{eq:1-bit_mlp_01} as
\[
\begin{aligned}
	\sup_{\boldsymbol{x} \in [-1,1]^H} \| f(\boldsymbol{x}) - f_{\textrm{1-bit}}(\boldsymbol{x}) \|
	&\geq \max_{\boldsymbol{x} \in \{-1,0,1\}^H} | f(\boldsymbol{x}) - f_{\textrm{1-bit}}(\boldsymbol{x}) | \\
	&= \max_{x_2, \dots, x_N \in \{-1,0,1\}^{H-1}}  | 1 - f_{\textrm{1-bit}}(\boldsymbol{x}) | + | 0 - f_{\textrm{1-bit}}(\boldsymbol{x}) | \\
	&\geq \max_{x_i \in \{-1,0,1\}^H}  | 1 - f_{\textrm{1-bit}}(\boldsymbol{x}) | + | f_{\textrm{1-bit}}(\boldsymbol{x}) | \\
	&\geq 1 \ ,
\end{aligned}
\]
where $i \in \{2,\dots, H\}$. The lower bound completes the proof. \hfill$\square$

\section{Full Proof of Theorem~\ref{thm:rate_att}} \label{app:thm_rate_att}
We take the straightforward computations on the approximation gap within the 2-norm.
\begin{equation}  \label{eq:att_degradation_total}
	\frac{1}{N_h} \left\| f_{\theta}( \boldsymbol{x} ) - f_{\hat{\theta}} (\boldsymbol{x}) \right\|_2 
	\leq  
	\left\| \rho (\boldsymbol{x}) \mathbf{W}_{\!\rho} \boldsymbol{x} - \hat{\rho} (\boldsymbol{x}) \hat{\mathbf{W}}_{\!\rho} \boldsymbol{x} \right\|_2 
	+ 
	\left\| \rho (\boldsymbol{x}) \boldsymbol{b}_\rho - \hat{\rho} (\boldsymbol{x}) \hat{\boldsymbol{b}}_\rho \right\|_2
	+ 
	\left\| \boldsymbol{b}_\text{o}- \hat{\boldsymbol{b}}_\text{o} \right\|_2 \ .
\end{equation}
Here, we ignore the superscript $(i)$ for simplicity. Specifically, we further bound each term of the right side of Eq.~\eqref{eq:att_degradation_total} as follows
\[
\begin{aligned}
	\left\| \rho (\boldsymbol{x}) \mathbf{W}_{\!\rho} \boldsymbol{x} - \hat{\rho} (\boldsymbol{x}) \hat{\mathbf{W}}_{\!\rho} \boldsymbol{x} \right\|_2 &\leq
	\left\| \rho (\boldsymbol{x}) \mathbf{W}_{\!\rho} \boldsymbol{x} - \rho^{(i)} (\boldsymbol{x}) \hat{\mathbf{W}}_{\!\rho} \boldsymbol{x} \right\|_2 
	+
	\left\| \rho (\boldsymbol{x}) \hat{\mathbf{W}}_{\!\rho} \boldsymbol{x} - \hat{\rho} (\boldsymbol{x}) \hat{\mathbf{W}}_{\!\rho} \boldsymbol{x} \right\|_2 \\
	&\leq
	\left\| \rho (\boldsymbol{x}) \right\|_2
	\left\|\mathbf{W}_{\!\rho} - \hat{\mathbf{W}}_{\!\rho}  \right\|_2
	\left\| \boldsymbol{x} \right\|_2
	+
	\left\| \rho (\boldsymbol{x}) -  \hat{\rho} (\boldsymbol{x}) \right\|_2 
	\left\| \hat{\mathbf{W}}_{\!\rho} \right\|_2
	\left\| \boldsymbol{x} \right\|_2  
\end{aligned}
\]
and
\[
\begin{aligned}
	\left\| \rho (\boldsymbol{x}) \boldsymbol{b}_\rho - \hat{\rho} (\boldsymbol{x}) \hat{\boldsymbol{b}}_\rho \right\|_2 
	&\leq
	\left\| \rho (\boldsymbol{x}) \boldsymbol{b}_\rho - \rho (\boldsymbol{x}) \hat{\boldsymbol{b}}_\rho \right\|_2
	+
	\left\| \rho (\boldsymbol{x}) \hat{\boldsymbol{b}}_\rho - \hat{\rho} (\boldsymbol{x}) \hat{\boldsymbol{b}}_\rho \right\|_2 \\
	&\leq
	\left\| \rho (\boldsymbol{x}) \right\|_2
	\left\| \boldsymbol{b}_\rho - \hat{\boldsymbol{b}}_\rho \right\|_2
	+
	\left\| \rho (\boldsymbol{x}) -  \hat{\rho} (\boldsymbol{x}) \right\|_2 
	\left\| \hat{\boldsymbol{b}}_\rho \right\|_2 \ . \\
\end{aligned}
\]
Since the ReLU activation is a 1-Lipschitz function, one has
\[
\begin{aligned}
	\left\| \rho (\boldsymbol{x}) \right\|_2
	&=
	\left\| \tau\left[ \left(\mathbf{W}_{\!\text{k}} \boldsymbol{x} + \boldsymbol{b}_\text{k} \right)^\top \left( \mathbf{W}_{\!\text{q}} \boldsymbol{x} + \boldsymbol{b}_\text{q} \right) \right] \right \| \\
	&\leq
	\left\| \left(\mathbf{W}_{\!\text{k}} \boldsymbol{x} + \boldsymbol{b}_\text{k} \right)^\top \left( \mathbf{W}_{\!\text{q}} \boldsymbol{x} + \boldsymbol{b}_\text{q} \right) \right\| \\
	& \leq 
	\left\| \boldsymbol{x}^{\top} \mathbf{W}_{\!\text{k}}^\top \mathbf{W}_{\!\text{q}} \boldsymbol{x} \right\|_2
	+ 
	\left\| \boldsymbol{x}^{\top} \mathbf{W}_{\!\text{k}}^\top \boldsymbol{b}_\text{q} \right\|_2 
	+
	\left\| \boldsymbol{b}_\text{k}^\top \mathbf{W}_{\!\text{q}} \boldsymbol{x} \right\|_2
	+
	\left\| \boldsymbol{b}_\text{k}^\top \boldsymbol{b}_\text{q} \right\|_2 \\
	&\leq
	\left\| \boldsymbol{x} \right\|_2^2
	\left\| \mathbf{W}_{\!\text{k}} \right\|_2
	\left\| \mathbf{W}_{\!\text{q}} \right\|_2
	+
	\left\| \boldsymbol{x} \right\|_2
	\left\| \mathbf{W}_{\!\text{k}} \right\|_2
	\left\| \boldsymbol{b}_{\text{q}} \right\|_2
	+
	\left\| \boldsymbol{b}_{\text{k}} \right\|_2
	\left\| \mathbf{W}_{\!\text{q}} \right\|_2
	\left\| \boldsymbol{x} \right\|_2
	+
	\left\| \boldsymbol{b}_{\text{k}} \right\|_2
	\left\| \boldsymbol{b}_{\text{q}} \right\|_2
\end{aligned}
\]
and
\[
\begin{aligned}
	\left\| \rho (\boldsymbol{x}) - \hat{\rho}  (\boldsymbol{x}) \right\|_2
	&= 
	\left\| \tau\left[ \left(\mathbf{W}_{\!\text{k}} \boldsymbol{x} + \boldsymbol{b}_\text{k} \right)^\top \left( \mathbf{W}_{\!\text{q}} \boldsymbol{x} + \boldsymbol{b}_\text{q} \right) \right] 
	-
	\tau\left[ \left(\hat{\mathbf{W}}_{\!\text{k}} \boldsymbol{x} + \hat{\boldsymbol{b}}_\text{k} \right)^\top \left( \hat{\mathbf{W}}_{\!\text{q}} \boldsymbol{x} + \hat{\boldsymbol{b}}_\text{q} \right) \right]
	\right \| \\
	&\leq
	\left\|  \left(\mathbf{W}_{\!\text{k}} \boldsymbol{x} + \boldsymbol{b}_\text{k} \right)^\top \left( \mathbf{W}_{\!\text{q}} \boldsymbol{x} + \boldsymbol{b}_\text{q} \right)
	-
	\left(\hat{\mathbf{W}}_{\!\text{k}} \boldsymbol{x} + \hat{\boldsymbol{b}}_\text{k} \right)^\top \left( \hat{\mathbf{W}}_{\!\text{q}} \boldsymbol{x} + \hat{\boldsymbol{b}}_\text{q} \right)
	\right \| \\
	&\leq
	\left\| \boldsymbol{x}^{\top} \mathbf{W}_{\!\text{k}}^\top \mathbf{W}_{\!\text{q}} \boldsymbol{x} - \boldsymbol{x}^{\top} \mathbf{W}_{\!\text{k}}^\top \hat{\mathbf{W}}_{\!\text{q}} \boldsymbol{x} \right\|_2
	+
	\left\| \boldsymbol{x}^{\top} \mathbf{W}_{\!\text{k}}^\top \hat{\mathbf{W}}_{\!\text{q}} \boldsymbol{x} -
	\boldsymbol{x}^{\top} \hat{\mathbf{W}}_{\!\text{k}}^\top \hat{\mathbf{W}}_{\!\text{q}} \boldsymbol{x} \right\|_2
	\\
	&\quad + 
	\left\| \boldsymbol{x}^{\top} \mathbf{W}_{\!\text{k}}^\top \boldsymbol{b}_\text{q} -\boldsymbol{x}^{\top} \mathbf{W}_{\!\text{k}}^\top \hat{\boldsymbol{b}}_\text{q} \right\|_2 
	+ 
	\left\| \boldsymbol{x}^{\top} \mathbf{W}_{\!\text{k}}^\top \hat{\boldsymbol{b}}_\text{q} -\boldsymbol{x}^{\top} \hat{\mathbf{W}}_{\!\text{k}}^\top \hat{\boldsymbol{b}}_\text{q} \right\|_2 \\
	&\quad +
	\left\| \boldsymbol{b}_\text{k}^\top \mathbf{W}_{\!\text{q}} \boldsymbol{x} - \boldsymbol{b}_\text{k}^\top \hat{\mathbf{W}}_{\!\text{q}} \boldsymbol{x} \right\|_2
	+ 
	\left\| \boldsymbol{b}_\text{k}^\top \hat{\mathbf{W}}_{\!\text{q}} \boldsymbol{x} - \hat{\boldsymbol{b}}_\text{k}^\top \hat{\mathbf{W}}_{\!\text{q}} \boldsymbol{x} \right\|_2 \\
	&\quad +
	\left\| \boldsymbol{b}_\text{k}^\top \boldsymbol{b}_\text{q} - \boldsymbol{b}_\text{k}^\top \hat{\boldsymbol{b}}_\text{q} \right\|_2
	+
	\left\| \boldsymbol{b}_\text{k}^\top \hat{\boldsymbol{b}}_\text{q} - \hat{\boldsymbol{b}}_\text{k}^\top \hat{\boldsymbol{b}}_\text{q} \right\|_2 \\
	&\leq 
	\left\| \boldsymbol{x} \right\|_2
	\left\| \mathbf{W}_{\!\text{k}} \right\|_2 
	\left\| \mathbf{W}_{\!\text{q}} - \hat{\mathbf{W}}_{\!\text{q}} \right\|_2 
	\left\| \boldsymbol{x} \right\|_2
	+ 
	\left\| \boldsymbol{x} \right\|_2
	\left\| \mathbf{W}_{\!\text{k}}-\hat{\mathbf{W}}_{\!\text{k}} \right\|_2
	\left\| \hat{\mathbf{W}}_{\!\text{q}} \right\|_2
	\left\| \boldsymbol{x} \right\|_2 \\
	&\quad +
	\left\| \boldsymbol{x} \right\|_2
	\left\| \mathbf{W}_{\!\text{k}} \right\|_2 
	\left\| \boldsymbol{b}_\text{q} - \hat{\boldsymbol{b}}_\text{q} \right\|_2
	+
	\left\| \boldsymbol{x} \right\|_2
	\left\| \mathbf{W}_{\!\text{k}} - \hat{\mathbf{W}}_{\!\text{k}} \right\|_2 
	\left\| \hat{\boldsymbol{b}}_\text{q} \right\|_2 \\
	&\quad +
	\left\| \boldsymbol{b}_\text{k} \right\|_2
	\left\| \mathbf{W}_{\!\text{q}} - \hat{\mathbf{W}}_{\!\text{q}} \right\|_2
	\left\| \boldsymbol{x} \right\|_2
	+
	\left\| \boldsymbol{b}_\text{k} - \hat{\boldsymbol{b}}_\text{k} \right\|_2
	\left\| \hat{\mathbf{W}}_{\!\text{q}} \right\|_2
	\left\| \boldsymbol{x} \right\|_2 \\
	&\quad +
	\left\| \boldsymbol{b}_\text{k} \right\|_2 
	\left\| \boldsymbol{b}_\text{q} - \hat{\boldsymbol{b}}_\text{q} \right\|_2
	+
	\left\| \boldsymbol{b}_\text{k} - \hat{\boldsymbol{b}}_\text{k} \right\|_2
	\left\| \hat{\boldsymbol{b}}_\text{q} \right\|_2 \ .
\end{aligned}
\]

Next, we bound the weight norms and the quantization gap. It is observed that $| \theta | \leq (n-1) $ for $\theta \in \mathcal{N}_n$; thus, we have
\begin{equation}  \label{eq:att_theta_bound}
	\left\{ ~\begin{aligned}
		&\| \mathbf{W}_\text{o}^{(i)} \|_2 \leq \sqrt{H N_v} \ (n-1) \ , \quad \| \boldsymbol{b}_\text{o}^{(i)} \|_2 \leq \sqrt{H} \ (n-1) \ , \\
		&\| \mathbf{W}_\text{v}^{(i)} \|_2 \leq \sqrt{N_v N} \ (n-1) \ , \quad \| \boldsymbol{b}_\text{v}^{(i)} \|_2 \leq \sqrt{N_v} \ (n-1) \ , \\
		& \| \mathbf{W}_\text{k}^{(i)} \|_2 \leq \sqrt{N_k  N} \ (n-1) \ , \quad \| \boldsymbol{b}_\text{k}^{(i)} \|_2 \leq \sqrt{N_k} \ (n-1) \ , \\
		& \| \mathbf{W}_\text{q}^{(i)} \|_2 \leq \sqrt{N_k  N} \ (n-1) \ , \quad \| \boldsymbol{b}_\text{q}^{(i)} \|_2 \leq \sqrt{N_k} \ (n-1) \ , \\
		& \left\| \mathbf{W}_{\!\rho}^{(i)} \right\|_2 \leq \left\| \mathbf{W}_{\!\text{o}}^{(i)} \right\|_2 \left\| \mathbf{W}_{\! \text{v}}^{(i)} \right\|_2 \leq \sqrt{HN} N_v (n-1)^2 \ , \\
		& \left\| \boldsymbol{b}_\rho^{(i)} \right\|_2
		\leq 
		\left\| \mathbf{W}_\text{o}^{(i)} \right\|_2
		\left\| \boldsymbol{b}_\text{v}^{(i)} \right\|_2 
		\leq \sqrt{H} N_v (n-1)^2 
	\end{aligned} \right.
\end{equation} 
and
\begin{equation}  \label{eq:att_theta_bound_gap}
	\left\{ ~\begin{aligned}
		&\| \mathbf{W}_\text{o}^{(i)} - \hat{\mathbf{W}}_\text{o}^{(i)} \|_2 \leq {H N_v} R(\theta, \hat{\theta}) \ , \quad 
		\| \boldsymbol{b}_\text{o}^{(i)} - \hat{\boldsymbol{b}}_\text{o}^{(i)} \|_2 \leq \sqrt{H} R(\theta, \hat{\theta}) \ , \\
		&\| \mathbf{W}_\text{v}^{(i)} - \hat{\mathbf{W}}_\text{v}^{(i)} \|_2 \leq N_v N R(\theta, \hat{\theta}) \ , \quad 
		\| \boldsymbol{b}_\text{v}^{(i)} - \hat{\boldsymbol{b}}_\text{v}^{(i)} \|_2 \leq \sqrt{N_v} R(\theta, \hat{\theta}) \ , \\
		& \| \mathbf{W}_\text{k}^{(i)} - \hat{\mathbf{W}}_\text{k}^{(i)} \|_2 \leq N_k N R(\theta, \hat{\theta}) \ , \quad 
		\| \boldsymbol{b}_\text{k}^{(i)} - \hat{\boldsymbol{b}}_\text{k}^{(i)} \|_2 \leq \sqrt{N_k} R(\theta, \hat{\theta}) \ , \\
		& \| \mathbf{W}_\text{q}^{(i)} - \hat{\mathbf{W}}_\text{q}^{(i)} \|_2 \leq N_k N R(\theta, \hat{\theta}) \ , \quad 
		\| \boldsymbol{b}_\text{q}^{(i)} - \hat{\boldsymbol{b}}_\text{q}^{(i)} \|_2 \leq \sqrt{N_k} R(\theta, \hat{\theta}) \ , \\
		& \left\| \mathbf{W}_{\!\rho}^{(i)} - \hat{\mathbf{W}}_{\!\rho}^{(i)} \right\|_2 \leq \left\| \mathbf{W}_{\!\text{o}}^{(i)} \right\|_2 \left\| \mathbf{W}_{\!\text{v}}^{(i)} - \hat{\mathbf{W}}_{\!\text{v}}^{(i)} \right\|_2 
		+
		\left\| \mathbf{W}_{\!\text{o}}^{(i)} - \hat{\mathbf{W}}_{\!\text{o}}^{(i)} \right\|_2
		\left\| \hat{\mathbf{W}}_{\!\text{v}}^{(i)} \right\|_2 \\
		&\qquad \qquad \qquad\quad~ \leq \left( \sqrt{HN^2} + \sqrt{H^2N} \right) N_v^{3/2} (n-1) R(\theta, \hat{\theta})  \\
		&\left\| \boldsymbol{b}_\rho^{(i)} - \hat{\boldsymbol{b}}_\rho^{(i)} \right\|_2
		\leq 
		\left\| \mathbf{W}_\text{o}^{(i)} \right\|_2
		\left\| \boldsymbol{b}_\text{v}^{(i)} - \hat{\boldsymbol{b}}_\text{v}^{(i)} \right\|_2
		+
		\left\| \mathbf{W}_\text{o}^{(i)} - \hat{\mathbf{W}}_\text{o}^{(i)} \right\|_2
		\left\| \hat{\boldsymbol{b}}_\text{v}^{(i)} \right\|_2 \\
		&\qquad \qquad \qquad \leq \left( H^{1/2}N_v + HN_v^{3/2} \right) (n-1) R(\theta, \hat{\theta}) \ ,
	\end{aligned} \right.
\end{equation} 
for $i \in [N_h]$, where $R(\theta, \hat{\theta})$ denotes the maximum of $| \theta - \hat{\theta} |$, or formally $R(\theta, \hat{\theta}) = \max_{\theta,\hat{\theta}} | \theta - \hat{\theta} |$. 

By substituting Eqs.~\eqref{eq:att_theta_bound} and~\eqref{eq:att_theta_bound_gap} into the above inequality, we have
\[
\left\{~ \begin{aligned}
	& \left\| \rho (\boldsymbol{x}) \right\|_2
	\leq
	N_k N_k \ (n-1)^2 \left( \sqrt{N} \ \| \boldsymbol{x} \|_2 + 1 \right)^2 \ , \\
	&\left\| \rho (\boldsymbol{x}) - \hat{\rho}  (\boldsymbol{x}) \right\|_2
	\leq
	N_k \left[ N^{\frac{3}{2}} N_k \| \boldsymbol{x} \|_2^2
	+
	\left( 2 N_k^{\frac{1}{2}} N + 2N^{\frac{1}{2}} \right)
	\| \boldsymbol{x} \|_2
	+ 2
	\right]  (n-1) \ R(\theta, \hat{\theta}) \ .
\end{aligned} \right.
\]
Further, Eq.~\eqref{eq:att_degradation_total} becomes
\[
\left\| f_{\theta}( \boldsymbol{x} ) - f_{\hat{\theta}} (\boldsymbol{x}) \right\|_2 
\leq  
N_h \left( 
\Delta_3 \| \boldsymbol{x} \|_2^3 +
\Delta_2 \| \boldsymbol{x} \|_2^2 + \Delta_1 \| \boldsymbol{x} \|_2 + \Delta_0
\right) R(\theta, \hat{\theta}) \ ,
\]
where
\[
\begin{aligned}
	& \Delta_0 = \mathcal{O} \left( H N N_k N_v^{\frac{3}{2}} n^3 \right) \ , \quad
	\Delta_1 = \mathcal{O} \left( H N^{\frac{3}{2}} N_k^{\frac{3}{2}} N_v^{\frac{3}{2}} n^3 \right) \ , \\
	& \Delta_2 = \mathcal{O} \left( H N^2 N_k^{\frac{3}{2}} N_v^{\frac{3}{2}} n^3 \right) \ ,\quad
	\Delta_3 = \mathcal{O} \left( H^{\frac{1}{2}} N^2 N_k^{\frac{3}{2}} N_v n^3 \right) \ .
\end{aligned}
\]
Let $\mu$ be a Lebesgue measure defined on $K_\text{in}$. Thus, we have
\begin{equation}  \label{eq:l2linfty_att}
	\left\{~ \begin{aligned} 
		&\left[ \int_{K_\text{in}} \left\|  f_{\theta}( \boldsymbol{x} ) - f_{\hat{\theta}} (\boldsymbol{x}) \right\|_2^2 \dif \mu(\boldsymbol{x}) \right]^{1/2} \\
		&\qquad\leq 
		C_2 N_h	\left[ \int_{K_\text{in}} \left[ \left( 
		\Delta_3 \| \boldsymbol{x} \|_2^3 +
		\Delta_2 \| \boldsymbol{x} \|_2^2 + \Delta_1 \| \boldsymbol{x} \|_2 + \Delta_0 \right) \right]^2 \dif \mu(\boldsymbol{x}) \right]^{1/2} R(\theta, \hat{\theta}) \ , \\
		&\esssup_{\boldsymbol{x} \in K_\text{in}}  \left\|  f_{\theta}( \boldsymbol{x} ) - f_{\hat{\theta}} (\boldsymbol{x}) \right\|_\infty \\
		&\qquad\leq 
		C_{\infty} N_h \esssup_{\boldsymbol{x} \in K_\text{in}} \left( 
		\Delta_3 \| \boldsymbol{x} \|_2^3 +
		\Delta_2 \| \boldsymbol{x} \|_2^2 + \Delta_1 \| \boldsymbol{x} \|_2 + \Delta_0
		\right) R(\theta, \hat{\theta}) \ . \\
	\end{aligned} \right.  
\end{equation}
Provided $K_\text{in} \subseteq [-D, D]^N$ where $D>0$, we have $\| \boldsymbol{x} \|_2 \leq \! \sqrt{N} D$ and $\| \boldsymbol{x} \|_\infty \leq D$. Further, one has
\begin{equation}  \label{eq:useful_4-1}
	\left[ \int_{K_\text{in}} \left\| \boldsymbol{x} \right\|_2^a  \dif \mu(\boldsymbol{x}) \right]^{1/2} \leq N^{\frac{a}{4}} D^{\frac{a}{2}} \ \mu^{\frac{a}{2}}(K_\text{in}) 
	\quad\text{and}\quad 
	\esssup_{\boldsymbol{x} \in K_\text{in}} \| \boldsymbol{x} \|^b_\infty \leq  D^b \ ,
\end{equation}
for $a \in [6]$ and $b \in [3]$. By substituting Eq.~\eqref{eq:useful_4-1} into Eq.~\eqref{eq:l2linfty_att}, we have
\[
\left\{~ \begin{aligned} 
	&\left[ \int_{K_\text{in}} \left\| f_{\theta}( \boldsymbol{x} ) - f_{\hat{\theta}} (\boldsymbol{x}) \right\|_2^2 \dif \mu(\boldsymbol{x}) \right]^{1/2} 
	\leq  
	\mathcal{O} \left( D^3 H N^{\frac{7}{2}} N_h N_k^{\frac{3}{2}} N_v^{\frac{3}{2}} n^3 \mu^3(K_\text{in}) \right) R(\theta, \hat{\theta}) 
	\ , \\
	&\esssup_{\boldsymbol{x} \in K_\text{in}}  \left\|  f_{\theta}( \boldsymbol{x} ) - f_{\hat{\theta}} (\boldsymbol{x}) \right\|_\infty 
	\leq 
	\mathcal{O} \left( D^3 H N^2 N_h N_k^{\frac{3}{2}} N_v^{\frac{3}{2}} n^3\right) R(\theta, \hat{\theta}) 
	\ .
\end{aligned} \right.  
\]
By setting
\[
\delta_1 = \mathcal{O} \left( \frac{\epsilon}{ D^3 H N^2 N_h N_k^{\frac{3}{2}} N_v^{\frac{3}{2}} n^3 } \right) \ , \quad
\delta_2 = \mathcal{O} \left( \frac{\epsilon}{ D^3 H N^{\frac{7}{2}} N_h N_k^{\frac{3}{2}} N_v^{\frac{3}{2}} n^3 \mu^3(K_\text{in}) } \right)   \ , 
\]
and $\hat{\theta} = Q_n(\theta)$, the followings hold
\[
\left\{~ \begin{aligned}
	& \frac{R(\theta, \hat{\theta})}{n-1} \leq \delta_1 
	\Rightarrow
	\left\|  f_{\theta}( \boldsymbol{x} ) - f_{\hat{\theta}} (\boldsymbol{x}) \right\|_{L^\infty(K_\text{in},\mathbb{R}^H)} \leq \epsilon  \ , \\
	& \frac{R(\theta, \hat{\theta})}{n-1} \leq \delta_2 
	\Rightarrow
	\left\|  f_{\theta}( \boldsymbol{x} ) - f_{\hat{\theta}} (\boldsymbol{x}) \right\|_{L^2(K_\text{in},\mathbb{R}^H)} \leq \epsilon \ . \\
\end{aligned} \right.
\]
We can finish the proof of Theorem~\ref{thm:rate_att}. $\hfill\square$

\section{Full Proof of Theorem~\ref{thm:rate_mlp}} \label{app:thm_rate_mlp}
It is evident that ReLU is a 1-Lipschitz activation function. For any $l \in [L]$, the approximation effects led by the connection parameter of the $l$-th layer becomes
\[
\begin{aligned}
\left\| f( \boldsymbol{x}; \mathbf{W}^{(l)} ) - f(\boldsymbol{x}; \mathbf{\hat{W}}^{(l)}) \right\|_2 
& \leq \left\| \mathbf{W}^{(L)} \boldsymbol{h}^{(L-1)} - \mathbf{W}^{(L)} \boldsymbol{\hat{h}}^{(L-1)} \right\|_2 \\
& \leq \left\| \mathbf{W}^{(L)} \right\|_2 ~\left\| \boldsymbol{h}^{(L-1)} - \boldsymbol{\hat{h}}^{(L-1)} \right\|_2 \\
& \leq \left( \prod_{l+1}^{L} \left\| \mathbf{W}^{(k)} \right\|_2 \right)~ \left\| \boldsymbol{h}^{(l)} - \boldsymbol{\hat{h}}^{(l)} \right\|_2 \\
& \leq \left( \prod_{k=l+1}^{L} \left\| \mathbf{W}^{(k)} \right\|_2 \right)~ \left\| \mathbf{W}^{(l)} \boldsymbol{h}^{(l-1)} - \mathbf{\hat{W}}^{(l)} \boldsymbol{h}^{(l-1)} \right\|_2 \\
& \leq \left( \prod_{k=l+1}^{L} \left\| \mathbf{W}^{(k)} \right\|_2 \right)~ \left\| \mathbf{W}^{(l)} - \mathbf{\hat{W}}^{(l)} \right\|_2 ~\left\| \boldsymbol{h}^{(l-1)} \right\|_2 \ .
\end{aligned}
\]
The above inequalities also hold for $\boldsymbol{b}^{(l)}$. Further, we consider the approximation effect led by a collection of connection parameters, that is, 
\begin{equation}  \label{eq:ftheta}
\begin{aligned}
	\left\| f_{\theta}( \boldsymbol{x} ) - f_{\hat{\theta}} (\boldsymbol{x}) \right\|_2 
	& \leq \left\| (  \mathbf{W}^{(L)} \boldsymbol{h}^{(L-1)} + \boldsymbol{b}^{(L)} ) - ( \mathbf{\hat{W}}^{(L)} \boldsymbol{\hat{h}}^{(L-1)} + \boldsymbol{\hat{b}}^{(L)} ) \right\|_2  \\
	& \leq \sum_{l=1}^L \left( \prod_{k=l+1}^L \left\| \mathbf{W}^{(k)} \right\|_2 \right) \Bigg( \left\| \mathbf{W}^{(l)} - \mathbf{\hat{W}}^{(l)} \right\|_2 \left\| \boldsymbol{\hat{h}}^{(l-1)} \right\|_2 + \left\| \boldsymbol{b}^{(l)} - \boldsymbol{\hat{b}}^{(l)} \right\|_2 \Bigg)  \ ,
\end{aligned}    
\end{equation}
in which
\[
\left\| \boldsymbol{h}^{(l)} \right\|_2 \leq \left( \prod_{k=1}^l \left\| \mathbf{W}^{(k)} \right\|_2 \right) \left\| \boldsymbol{x} \right\|_2 
+  \sum_{k=1}^l \left( \prod_{h=k+1}^{l} \left\| \mathbf{W}^{(h)} \right\|_2 \right) \left\| \boldsymbol{b}^{(k)} \right\|_2    
\]
and
\[
\left\| f_{\theta}( \boldsymbol{x} ) \right\|_2  \leq  \left( \prod_{l=1}^L \left\| \mathbf{W}^{(l)} \right\|_2 \right) \left\| \boldsymbol{x} \right\|_2 
+ \sum_{l=1}^L \left( \prod_{k=l+1}^L \left\| \mathbf{W}^{(k)} \right\|_2 \right) \left\| \boldsymbol{b}^{(l)} \right\|_2 \ .    
\]
It is observed that $| \theta | \leq (n-1) $ for $\theta \in \mathcal{N}_n$; thus, we have 
\[
\| \mathbf{W}^{(l)} \|_2 \leq N_w (n-1)
\quad\text{and}\quad
\| \mathbf{b}^{(l)} \|_2 \leq \sqrt{N_w} (n-1)
\quad\text{for}\quad
l \in [L] \ .
\]
Let $R(\theta, \hat{\theta})$ denote the maximum of $| \theta - \hat{\theta} |$, or formally $R(\theta, \hat{\theta}) = \max_{\theta,\hat{\theta}} | \theta - \hat{\theta} |$. Then we have 
\[
\| \mathbf{W}^{(l)} - \mathbf{\hat{W}}^{(l)} \|_2 \leq N_w R(\theta, \hat{\theta})
\quad\text{and}\quad
\| \boldsymbol{b}^{(l)} - \boldsymbol{\hat{b}}^{(l)}\|_2 \leq \sqrt{N_w} R(\theta, \hat{\theta})
\quad\text{for}\quad
l \in [L] \ .
\]
Thus, Eq.~\eqref{eq:ftheta} becomes
\[
\begin{aligned} 
\left\| f_{\theta}( \boldsymbol{x} ) - f_{\hat{\theta}} (\boldsymbol{x}) \right\|_2
&\leq
\sum_{l=1}^L ( N_w n )^{L-l} \left[ ( N_w n )^{l-1} \left\| \boldsymbol{x} \right\|_2 + \sum_{k=0}^{l-1} ( N_w n )^{k} \right]  N_w R(\theta, \hat{\theta}) \\
&\leq \left\{  \left[ \sum_{l=1}^L ( N_w n )^{L-1} \right] \left\| \boldsymbol{x} \right\|_2 
+ \left[ \sum_{l=1}^L ( N_w n )^{L-l} \frac{(N_w n)^l - 1}{N_w n - 1} \right] \right\} N_w R(\theta, \hat{\theta})  \\
&\leq \left[ L ( N_w n )^{L-1} \left\| \boldsymbol{x} \right\|_2  + \frac{L( N_w n )^L}{ N_w n - 1} \frac{( N_w n )^L - 1}{( N_w n - 1 )^2} \right]  N_w R(\theta, \hat{\theta})  \ .
\end{aligned}  
\]
For convenience, we take a short notation
\[
\Delta(\boldsymbol{x}, N_w, L, n)
\overset{\mathrm{def}}{=} 
N_w \left[
L ( N_w n )^{L-1} \left\| \boldsymbol{x} \right\|_2  + \frac{L( N_w n )^L}{ N_w n - 1} \frac{( N_w n )^L - 1}{( N_w n - 1 )^2} \right] \ .    
\]
It is easy to obtain that $\Delta(\boldsymbol{x}, N_w, L, n) = \mathcal{O} ( L N_w^L n^{L-1} ) ( \| \boldsymbol{x} \|_2 + c )$, where $c$ is a universal constant. Let $\mu$ be a Lebesgue measure defined on $K_\text{in}$. Thus, we have
\begin{equation}  \label{eq:l2linfty}
\left\{ \begin{aligned} 
	\left[ \int_{K_\text{in}} \left\|  f_{\theta}( \boldsymbol{x} ) - f_{\hat{\theta}} (\boldsymbol{x}) \right\|_2^2 \dif \mu(\boldsymbol{x}) \right]^{1/2}
	&\leq 
	C_2 \left[ \int_{K_\text{in}} \Delta(\boldsymbol{x}, N_w, L, n)^2 R(\theta, \hat{\theta})^2  \dif \mu(\boldsymbol{x}) \right]^{1/2} \\
	&\leq \mathcal{O} \left( L N_w^L n^{L-1} \right) \left[ \int_{K_\text{h}} ( \left\| \boldsymbol{x} \right\|_2 + c )]^2 \dif \mu(\boldsymbol{x}) \right]^{1/2}  R(\theta, \hat{\theta})  \ , \\
	\esssup_{\boldsymbol{x} \in K_\text{h}}  \left\|  f_{\theta}( \boldsymbol{x} ) - f_{\hat{\theta}} (\boldsymbol{x}) \right\|_\infty 
	&\leq 
	C_{\infty} \esssup_{\boldsymbol{x} \in K_\text{h}} \left[  N_w \Delta(\boldsymbol{x}, N_w, L, n) R(\theta, \hat{\theta}) \right] \\
	&\leq 
	\mathcal{O} \left( L N_w^L n^{L-1} \right) \left( \esssup_{\boldsymbol{x} \in K_\text{h}} \| \boldsymbol{x} \|_\infty + c \right) R(\theta, \hat{\theta}) \ . \\
\end{aligned} \right.  
\end{equation}
Provided $K_\text{h} \subseteq [-D, D]^H$ where $D>0$, we have $\| \boldsymbol{x} \|_2 \leq \! \sqrt{H} D$ and $\| \boldsymbol{x} \|_\infty \leq D$. Further, one has
\begin{equation}  \label{eq:useful_3-1}
\left[ \int_{K_\text{h}} ( \left\| \boldsymbol{x} \right\|_2 + c )^2 \dif \mu(\boldsymbol{x}) \right]^{1/2} \leq \sqrt{H} D \mu(K_\text{h}) 
\quad\text{and}\quad
\esssup_{\boldsymbol{x} \in K_\text{h}} \| \boldsymbol{x} \|_\infty \leq  D \ .  
\end{equation}
By substituting Eqs.~\eqref{eq:useful_3-1} into Eq.~\eqref{eq:l2linfty}, we have
\begin{equation}  \label{eq:l_2-l_infty}
\left\{~ \begin{aligned} 
	\left\|  f_{\theta}( \boldsymbol{x} ) - f_{\hat{\theta}} (\boldsymbol{x})  \right\|_{L_\mu^2(K_\text{h},\mathbb{R}^M)}
	& \leq 
	\mathcal{O} \left( L N_w^L n^{L-1} \sqrt{H} D \mu(K_\text{h}) \right) R(\theta, \hat{\theta}) \ , \\
	\left\| f_{\theta}( \boldsymbol{x} ) - f_{\hat{\theta}} (\boldsymbol{x})  \right\|_{L_\mu^\infty(K_\text{h},\mathbb{R}^M)} 
	& \leq 
	\mathcal{O} \left( L N_w^L n^{L-1} D \right) R(\theta, \hat{\theta}) \ .
\end{aligned} \right.  
\end{equation}
By setting $\hat{\theta} = Q_n(\theta)$, the followings hold
\[
\left\{~ \begin{aligned}
& \frac{R(\theta, \hat{\theta})}{n-1} \leq \delta_1 = \mathcal{O} \left( \frac{\epsilon}{ L N_w^L n^L D } \right) 
\Rightarrow
\left\|  f_{\theta}( \boldsymbol{x} ) - f_{\hat{\theta}} (\boldsymbol{x}) \right\|_{L^\infty(K_\text{h},\mathbb{R}^M)} \leq \epsilon \\
& \frac{R(\theta, \hat{\theta})}{n-1} \leq \delta_2 = \mathcal{O} \left( \frac{\epsilon}{ L N_w^L n^L \sqrt{H} D \mu(K_\text{h}) } \right)
\Rightarrow
\left\|  f_{\theta}( \boldsymbol{x} ) - f_{\hat{\theta}} (\boldsymbol{x}) \right\|_{L^2(K_\text{h},\mathbb{R}^M)} \leq \epsilon \ . \\
\end{aligned} \right.
\]
We can finish the proof of Theorem~\ref{thm:rate_mlp}. $\hfill\square$

\clearpage
\bibliographystyle{abbrvnat}
\bibliography{refs}

\end{document}